\documentclass[sn-mathphys]{sn-jnl}
\jyear{2022}
\theoremstyle{thmstyleone}%
\newtheorem{theorem}{Theorem}%  meant for continuous numbers
\newtheorem{proposition}[theorem]{Proposition}% 
\normalbaroutside
\theoremstyle{thmstyletwo}%
\newtheorem{example}{Example}%
\newtheorem{remark}{Remark}%
\theoremstyle{thmstylethree}%
\newtheorem{definition}{Definition}%
\graphicspath{ {./images/} }
\usepackage{setspace}

\usepackage{xurl}
\usepackage{amsmath,amsfonts}
\usepackage{textcomp}
\usepackage{enumitem}
\usepackage{algorithm}
\usepackage{algpseudocode}
\usepackage{mwe}
\usepackage{amsthm}
\newtheorem{corollary}{Corollary}
\usepackage{float}
\usepackage{placeins}  
\usepackage{hyperref}
\usepackage{tabularx}
\usepackage{comment}
\usepackage{anyfontsize}
\usepackage{mathtools,amssymb,amsmath,latexsym,faktor,kotex,stmaryrd}
\usepackage{varwidth}
\usepackage{graphicx}
\usepackage{tikz}
\usepackage{float}
\usepackage{afterpage}
\usetikzlibrary{shapes.geometric}
\usepackage{subcaption}
\usepackage{program}
\usepackage{verbatim}
\tikzset{
amp/.style = {regular polygon, regular polygon sides=3,
              draw, fill=white, text width=1em,
              inner sep=1mm, outer sep=0mm,
              shape border rotate=0},
amp1/.style = {regular polygon, regular polygon sides=3,
              draw, fill=white, text width=1em,
              inner sep=2mm, outer sep=0mm,
              shape border rotate=0},
amp1/.style = {regular polygon, regular polygon sides=3,
              draw, fill=white, text width=1em,
              inner sep=2mm, outer sep=0mm,
              shape border rotate=0},
amp2/.style = {regular polygon, regular polygon sides=3,
              draw, fill=white, text width=1em,
              inner sep=3.83mm, outer sep=0mm,
              shape border rotate=0},
amp3/.style = {regular polygon, regular polygon sides=3,
              draw, fill=white, text width=1em,
              inner sep=3.83mm, outer sep=0mm,
              shape border rotate=0},
amp4/.style = {regular polygon, regular polygon sides=3,
              draw, fill=white, text width=1em,
              inner sep=3.83mm, outer sep=0mm,
              shape border rotate=0},
amp5/.style = {regular polygon, regular polygon sides=3,
              draw, fill=white, text width=1em,
              inner sep=2.4mm, outer sep=0mm,
              shape border rotate=0}
}

\begin{document}
\title[WIF and WRCF]{Weighted  Isolation and Random Cut Forests for Anomaly Detection}

\author[1,2]{\fnm{Sijin} \sur{Yeom}}\email{yeomsijin@postech.ac.kr}
\author*[1,2,3]{\fnm{Jae-Hun} \sur{Jung}}\email{jung153@postech.ac.kr}

%%%%%%%%
\affil[1]{\orgdiv{Artificial Intelligence Graduate School},
\orgname{Pohang University of Science and Technology}, 
\orgaddress{\city{Pohang}, 
\postcode{37673},\country{Korea}}}

\affil[2]{\orgdiv{Mathematical Institute of Data Science},
\orgname{Pohang University of Science and Technology}, 
\orgaddress{\city{Pohang}, 
\postcode{37673},\country{Korea}}}

\affil[3]{\orgdiv{Department of Mathematics}, \orgname{Pohang University of Science and Technology}, \orgaddress{ \city{Pohang}, \postcode{37673},  \country{Korea}}}

\abstract{Random cut forest (RCF) algorithms have been developed for anomaly detection, particularly in time series data. The RCF algorithm is an improved version of the isolation forest (IF) algorithm. Unlike the IF algorithm, the RCF algorithm can determine whether real-time input contains an anomaly by inserting the input into the constructed tree network. Various RCF algorithms, including Robust RCF (RRCF), have been developed, where the cutting procedure is adaptively chosen probabilistically. The RRCF algorithm demonstrates better performance than the IF algorithm, as dimension cuts are decided based on the geometric range of the data, whereas the IF algorithm randomly chooses dimension cuts. However, the overall data structure is not considered in both IF and RRCF, given that split values are chosen randomly. In this paper, we propose new IF and RCF algorithms, referred to as the weighted IF (WIF) and weighted RCF (WRCF) algorithms, respectively. Their split values are determined by considering the density of the given data. To introduce the WIF and WRCF, we first present a new geometric measure, a \textit{density measure}, which is crucial for constructing the WIF and WRCF. We provide various mathematical properties of the density measure, accompanied by theorems that support and validate our claims through numerical examples.}


\maketitle
\section{Introduction}\label{introduction}
Anomaly detection is a crucial data mining procedure used in various applications such as fraud detection in finance, detection of network intrusion in web infrastructure, etc. \cite{Bartos2019, Goldstein_2015}. Anomaly detection is a detection algorithm for outliers that deviate from the normal structure of the given data set. 

Supervised anomaly detection algorithms use the labeled data, that is, the algorithms are constructed based on the known definition of `normal' and `abnormal' of the data points of the data. In this case, it is straightforward to detect anomalies when the algorithms are well-trained with balanced data sets. Semi-supervised anomaly detection algorithms can also be considered when the labeled samples are limited.  However, anomalies are intrinsically rare events, and, in most cases, anomaly samples are rare, thus sometimes it is impossible to label them a priori \cite{10.1371/journal.pone.0152173}. 
Thus the desired anomaly detection algorithms can acquire the notion of anomaly from the data structure without supervision or with semi-supervision. 

Such anomaly detection algorithms can be constructed either statistically or geometrically. 
Recently deep neural network algorithms have been actively developed for anomaly detection \cite{ruff2020deep, 9420659}. 
Unlike the neural network approaches, neighbor search approaches such as mining outliers using distance based on the $k$-th nearest neighbor \cite{peterson2009k, patrick1970generalized, dudani1976distance, gates1972reduced} and
density-based local outliers \cite{breunig2000lof, kriegel2009loop, he2003discovering} or the tree approaches such as the isolation forest (IF) \cite{4781136, 8888179} and robust random cut forest (RRCF) algorithms \cite{guha2016robust, bartos2019rrcf} are geometric. 

The geometric approaches based on the neighbor search using the Euclidean distance are computationally costly. The tree approaches are relatively less costly as they avoid the direct computation of pair-wise distances. The IF and RRCF algorithms partition the given data set forming the binary trees and defining the anomaly score based on the tree.  With the ordinary partitioning, however, after choosing a dimension cut, a split value is determined by only two values out of the whole data set. That is,  the overall shape of a given data set is not considered.

To consider the overall data structure and improve the existing IF and RRCF algorithms, we introduce a new and improved approach, termed the \textit{Weighted Isolation Forest} (WIF) and the \textit{Weighted Random Cut Forest} (WRCF) algorithms, respectively. In these algorithms, we determine the split values by taking into account the \textit{density} of the data set.

The paper is organized into the following sections. In Sections \ref{IsolationForest} and \ref{RobustRandomCutForest}, we review the IF and RRCF algorithms, respectively. In Section \ref{density measure}, we introduce the density measure and outline its various mathematical characteristics. Section \ref{Weighted Isolation Forest and Weighted Random Cut Forest} presents our proposed algorithms, namely, the WIF and WRCF. 
In Section \ref{Examples}, we provide numerical examples demonstrating that the proposed WIF and WRCF outperform their counterparts, IF and RRCF, especially in scenarios with high data density. The examples aim to illustrate the superior performance of our proposed algorithms.
Finally, Section \ref{Conclusion} offers a concise concluding remark, summarizing the key findings and contributions of the paper.

\section{Isolation Forest}
\label{IsolationForest}
We first explain the isolation forest algorithm introduced in  \cite{4781136}.\ In this section we formalize mathematical definitions and propositions necessary for the construction of the WIF and WRCF defined in Section \ref{Weighted Isolation Forest and Weighted Random Cut Forest}. The IF algorithm builds a tree by randomly cutting the \textit{bounding boxes} of the data set until all instances are \textit{isolated}. This algorithm doesn't require computing pairwise distances between data points, which can be computationally expensive, especially when the dimension of the data is large.

Given a data set, we create an isolation forest which consists of isolation trees. In an isolation tree, each external \textit{node} of the tree is assigned to an \textit{anomaly score}. The IF algorithm detects anomalies by calculating the anomaly score of each point; the closer the score is to 1, the higher the likelihood of it being an anomaly.

\subsection{Binary Trees}
When we create a binary tree from data sets in $\mathbb{R}^d$, we utilize the projection of data sets onto a specific axis. In this case, the projected data sets may contain duplicates. Hence, throughout the paper,  we assume that our data set $X\subseteq \mathbb{R}^d$ is a \textbf{finite multiset} of $n\ge 2$ elements.

An isolaton tree is a \textit{binary tree} constructed in a specific way. 
\begin{definition}
A \textbf{binary tree} $T$ of $X$ is a collection of nonempty subsets of $X$ such that 
\begin{enumerate}[label=\roman*)]
\item $X \in T$;
\item If $D \in T$ with $D \neq X$, then there exists
$ D'  \in T$ such that 
 \[ D \cup D' \in T  \; \text{ and  } \; D \cap D' =\varnothing.\] 
\end{enumerate}
\begin{itemize}
\item A \textbf{subtree} $T'$ in a binary tree $T$ of $X$   is a binary tree of $X' \subseteq X$ such that $ T' \subseteq T$.
\item An element of $T$ is called a \textbf{node} of $T$. 
\item The largest node, $X$, is called the \textbf{root node} of $T$.	
\end{itemize}
\end{definition}

\begin{definition}
A \textbf{graph} of a binary tree $T$ is an ordered pair $(V,E)$ such that
\begin{enumerate}[label=\roman*)]
\item $V=T$;
\item $E = \{ \{ D, D_{\text{p}} \}  \subseteq V \mid D \subseteq D_{\text{p}},  \;    D_{\text{p}} \setminus D \in  V\}$
\end{enumerate}
where $V$ is the set of vertices (or nodes) and $E$ is the set of edges.	
\end{definition}

\begin{definition}
Let $D,D'$ and $D_{\text{p}}$ be nodes in a binary tree $T$ where $D_\text{p}$ is a disjoint union of $D$ and $D'$.
\begin{itemize}
\item	The node $D_\text{p}$ is said to be a \textbf{parent} node of $D$ and $D'$.
\item The two nodes $D$ and $D'$ are called \textbf{children} nodes of $D_\text{p}$.
\item  We say $D'$ is the  \textbf{sibling} node of $D$ and vice versa.
\item A node that has no children nodes is an \textbf{external} (or a \textbf{leaf}) node 
in $T$.
\item A node that is not an external node is an \textbf{internal} node in $T$.
\item If every external node of $T$ is a singleton set, then $T$ is said to be \textbf{fully grown}.
\end{itemize}
	
\end{definition}

\begin{definition}
Let $T$ be a binary tree of $X$ and $D$ be a node in $T$.
\begin{itemize}
\item 
The \textbf{depth} of $D$, denoted by $d(D, X, T)$, is defined to be 
the number of edges from $D$ to the root node $X$ in the graph of $T$.
\item
We define the \textbf{height} of the tree $T$ by the maximum depth of nodes in $T$, that is, 
\[
\text{the height of $T$} = \max\{  d(D,X,T) \mid D \in T\}.
\]	
\end{itemize}

\end{definition}

\FloatBarrier

\subsection{Isolation Trees}
An isolation tree is obtained from
a ‘partitioning process' which cuts the \textit{bounding boxes} repeatedly.
We introduce the basic concept used for the isolation tree, i.e., the partitioning process, and explain how to create an isolation tree through it.

\begin{definition}\label{xq}
For $q=1,2,\dots,d$,
we denote the set of the $q$-th coordinate values in $X$ by
\begin{equation}\label{eq1}
  X_q= \{ x_q \mid (x_1,\dots,x_q,\dots,x_d) \in X\} \subseteq \mathbb{R}    
\end{equation}
which is the projection of $X$ onto the $q$-th axis.
We denote the length of each $X_q$ by

\begin{equation}\label{eq2}
l_q = \max X_q - \min X_q.
\end{equation}
We define the the \textbf{bounding box} of $X$ by
\[
B(X) = \prod\limits_{q=1}^d [\min X_q, \max X_q ]
\]
which is the smallest closed cube in $\mathbb{R}^d$ containing $X$.
\end{definition}

\noindent\textbf{Partitioning process for isolation trees}.\\
\noindent\textbf{Step 1.} Consider the equally distributed discrete random variable $Q$ of range $\{1,2,\dots,d\} $ such that:
\[ \text{Prob}^{\text{IF}}(Q=q)=\frac{1}{d} \]
for all $q=1,2,\dots,d$. Here the chosen $Q=q$ is called the \textbf{dimension cut}.
\;

\noindent\textbf{Step 2.}
Given $Q=q$, we now consider a random value $P|_{Q=q}$ which is uniformly distributed 
on the closed interval 
$[\min X_q, \max X_q]$. 
Here, the chosen $P|_{Q=q}=p$ is called the \textbf{split value}.

\noindent\textbf{Step 3.}
With the dimension cut $q \in \{1,2,\dots, d \}$ and the split value $p \in [\min X_q, \max X_q]$,
 we have  two outputs
\begin{displaymath}
\begin{split}
	  S_{1} & = \{ (x_1,\dots, x_q, \dots, x_d) \in X \mid x_q < p \}, \\
	  S_{2} & = \{ (x_1,\dots, x_q, \dots, x_d)  \in X \mid x_q \ge p \}
\end{split}
\end{displaymath}
where $\{S_1, S_2 \}$ is a partition of $X$.

To build an isolation tree of $X$, we continue the partitioning process  until either 
\begin{enumerate}[label = \roman*)]
\item both outputs become singleton sets, 
\item or, $\min X_q = \max X_q$ for some dimension cut $q$.
\end{enumerate}
\noindent At the end of the partitioning process, a collection of all outputs, including the first input $X$ 
forms a binary tree called an \textbf{isolation tree} of $X$.

\begin{definition}
A collection of isolation trees is called an \textbf{isolation forest}.
\end{definition}

\subsection{Anomaly Score}
In the IF algorithm, we use the anomaly score $s(x,X,F)$   \cite{4781136} where
 $x$ is a point in a data set $X$
and $F$ is an isolation forest.
Given a data set $X$ of $n$ elements, 
the maximum possible height of an isolation tree is $n-1$ and it grows 
in the order of $n$.
But 
the average height of an isolation tree, $c(n)$,  is  given by
\[
c(n)= 2(\ln(n-1)+\gamma) -\frac{2(n-1)}{n}
\]
which grows in the order of $\ln(n)$ \cite{Knuth1973} 
where $\gamma$ is the Euler–Mascheroni constant:
\[
\gamma = \lim\limits_{n \rightarrow \infty} 
\left( 
\sum\limits_{k=1}^n \frac{1}{k} - \ln n
\right)
\approx 0.5772156649.
\]
\begin{definition}Let $F$ be an isolation forest of $X$. 
The anomaly score $s(x,X,F)$ 
 is defined by \cite{4781136}
\[
s(x,X,F)= 2^{-\frac{\mathbb{E}(d(D_x,X,F))}{c(n)}} 
\]
where $D_x $ is  the external node containing $x$ and
\begin{displaymath}
\begin{split}
	\mathbb{E}(d(D_x,X,F) )&= 
\frac{1}{\vert F_x \vert }  \sum\limits_{T \in F_x} d(D_x,X,T)
\end{split}
\end{displaymath}
with $F_x= \{T \in F \mid D_x \in T \}$.
\end{definition}

\begin{remark} \cite{4781136}
\begin{enumerate}[label=\roman*)]
\item The anomaly score $s(x,X,F)$ is normalized in $(0,1]$.
\item The anomaly score $s(x,X,F)$ is a decreasing function with respect to $d(D_x,X,F)$.
\item If $s(x, X, F) \approx 0.5$ for every $x\in X$, then $d(D_x, X, F) \approx c(n)$, that is, the average depth of $x$ in the forest is close to the average height of the isolation trees in the forest. In this case, we may conclude that $X$ has no anomalies.
\item If $s(x,X,F)$ is close to 1, then $x$ is regarded as an anomaly.
\item If $s(x,X,F)$ is less than 0.5, then $x$ is regarded as a normal point.
\end{enumerate}
\end{remark}

\section{Robust Random Cut Forest}\label{RobustRandomCutForest}
In contrast to the IF, the RRCF consists of \textit{robust random cut trees} (RRCTs) and each external node of a random cut tree has an anomaly score called \textit{collusive displacement}.

\subsection{Robust Random Cut Trees}
The way of building an RRCT from a data set is the same as 
the way of building an isolation tree except for the random variable $Q$.
In an isolation tree, a random variable $Q$ has 
a uniform distribution that only depends on $d$.
In an RRCT, however, a random variable $Q$ depends on the 
range of each coordinate value of the data set as well as on $d$.

\vspace{0.3cm}

\noindent\textbf{Partitioning process for RRCTs.}\\
\noindent\textbf{Step 1.} Consider the discrete random variable $Q$  of range $\{1,2,\dots,d\}$ such that
\[ \text{Prob}^{\text{RRCF}}(Q=q)=\frac{l_q}{l_1+\dots+l_d} \]
for $q=1,2,\dots,d$. 
Here the definition of $l_q$ is given in Section \ref{IsolationForest}, specifically in Eq. (\ref{eq2}).

\noindent\textbf{Step 2.}\label{rrcfstep2}
Given the dimension cut $q$, we now consider a random value $P|_{Q=q}$ which is uniformly distributed 
on the closed interval 
$[\min X_q, \max X_q]$.

\noindent\textbf{Step 3.}
With the dimension cut $q \in \{1,2,\dots, d \}$ and  the split value $p \in [\min X_q, \max X_q ]$,
 we have a  partition of $X$:
\begin{displaymath}
\begin{split}
	  S_{1} & = \{ (x_1,\dots, x_q, \dots, x_d) \in X \mid x_q < p \}, \\
	  S_{2} & = \{ (x_1,\dots, x_q, \dots, x_d)  \in X \mid x_q \ge p \}.
\end{split}
\end{displaymath}

To build a robust random cut tree of $X$, we continue the partitioning process until 
both outputs become singleton sets.
\noindent At the end of the process, a collection of all outputs, including the first input $X$,
forms a binary tree called a \textbf{RRCT} of $X$.

\begin{remark}
\;
\begin{enumerate}[label=\roman*)]
	\item $\text{Prob}^{\text{RRCF}}(Q=q)$ is an increasing function with respect to the length $l_q$.
\item 
If $\min X_q= \max X_q $
 for some $q$,  then $l_q=0$ and thus $\text{Prob}^{\text{RRCF}}(Q=q)=0$.
Hence we only choose $Q=q$ such that $\min X_q < \max X_q $.
\item 
The partitioning process for RRCTs takes the set $X$ as an input and always  gives  us 
two mutually disjoint nonempty subsets $S_1, S_2$ of $X$ as outputs.
In other words, unlike isolation trees, RRCTs are always fully grown.
\end{enumerate}
\end{remark}

\begin{definition}
A collection of RRCTs is called a \textbf{robust random cut forest}.
\end{definition}

\subsection{Anomaly Score}
Let $T$ be a robust random cut tree of  $X$.
The sum of the depths of all nodes in $T$
\[
\vert M(T) \vert = \sum\limits_{D \in T} d(D,X,T)
\]
measures how completed the tree $T$ is, so, it is called the \textbf{model complexity} of $T$  \cite{guha2016robust}.

The \textbf{collusive displacement} of $x \in X$, denoted by $\textbf{CODISP}(x, X, T)$, is defined to be \cite{guha2016robust}
\[
\max\limits_{x \in D \subsetneq X} 
\left \{
\frac{1}{\vert D \vert} \sum\limits_{y \in X \setminus D}
 (d(\{y\},X,T) -d(\{y\},X\setminus D,T')) 
 \right \}
\]
where $T'$ is the tree of $X\setminus D$  such that 
\[
T' = \{M \setminus D \mid M \in T   \}.
\]
The graph of $T'$ is  obtained by deleting the subtree of $D$ in $T$  and by replacing the parent node $D_\text{p}$ of $D$ with 
the subtree of the sibling node $D_{\text{s}}$ of $D$ (Figs. \ref{fig1} and \ref{fig2}).
\begin{figure}[h]
\centering
\begin{tikzpicture}
[
level 1/.style = {sibling distance = 4cm},
level 2/.style = {sibling distance = 2cm},
level 3/.style = {sibling distance = 3cm}
]
\node  {$X$}
		child { node { $\vdots$} 
			child { node[blue]{ $D_{\text{p}}$}     
				child { node {$D$} 
					child { node[amp5] {}}}
				child { node {$D_{\text{s}}$}  
					child { node[red, amp5] {}}				
							 }}}
		child {node {}
			child { node[amp4] {}}}  ;
 \end{tikzpicture}
\caption{The graph of $T$}
\label{fig1}
\end{figure}

\begin{figure}[h]
\centering
\begin{tikzpicture} 
[
level 1/.style = {sibling distance = 4cm},
level 2/.style = {sibling distance = 2cm},
level 3/.style = {sibling distance = 3cm}
]
\node  {$X \setminus D$}
		child { node { $\vdots$} 
			child { node[blue] {$D_{\text{s}}$}  
					child { node[red, amp5] {}}				
							 }}
		child {node {}
			child { node[amp4] {}}}  ;
 \end{tikzpicture}
\caption{The graph of $T'$}
\label{fig2}
\end{figure}

The following Proposition \ref{prop1} was stated in \cite{guha2016robust} without proof.
Here we provide its proof.
\begin{proposition}\label{prop1}
Let $T$ be an RRCT of $X$ and $D$ be a node in $T$ with $D \neq X$ and $T'$ be the tree obtained by deleting the subtree of $D$ in $T$. 
Then 
\[
\sum\limits_{y \in X \setminus D}
 (d(\{y\},X,T) -d(\{y\},X\setminus D,T'))  = \vert D_{\text{s}} \vert
\]
where $D_\text{s}$ is the sibling node of $D$.
\end{proposition}

\begin{proof}
Since $T$ is a binary tree and $D \neq X$, $D$ has the parent node $D_\text{p}$ and the sibling node $D_\text{s}$.
Let $T'$ be the tree of $X \setminus D$ obtained from $T$ deleting the subtree of $D$.
Suppose $y \in X \setminus D$. 
If $y \notin D_{\text{s}}$, then  the number of edges from $\{y\}$ to the root node $X$ in the tree $T$ and 
the number of edges from $\{y\}$ to the root node $X \setminus D$ in the tree $T'$ are the same.
Thus 
 \[
d(\{y\},X,T) - d(\{y\},X\setminus D,T')=0
\]
for each $ y \in X \setminus (D \cup D_{\text{s}})$.

If $y \in D_{\text{s}}$, the number of edges in $T$ from $\{y\}$ to the root node 
is greater by 1 than those in $T'$, that is, 
the edge $\{D_{\text{p}}, D_{\text{s}} \}$ is included in $T$;
 \[
d(\{y\},X,T) -d(\{y\},X\setminus D,T')=1
\]
for each $ y \in  D_{\text{s}}$. Hence we have

\begin{displaymath}
\begin{split}
&\sum\limits_{y \in X \setminus D}
 (d(\{y\},X,T) -d(\{y\},X\setminus D,T')) \\
=&\sum\limits_{y \in D_{\text{s}} }
(d(\{y\},X,T) -d(\{y\},X\setminus D,T'))  \\
=& \vert D_\text{s} \vert.
\end{split}
\end{displaymath}
\end{proof}

\begin{corollary}
The \text{collusive displacement} of $x \in X$ can be calculated effectively  by 
\[
\textbf{CODISP}(x, X, T) =\max
\left\{
\frac{\vert D_{\text{s}} \vert}{\vert D \vert} \mid {x \in D \subsetneq X} 
 \right\}.
\]
\end{corollary}

\subsection{RRCF with Bagging}
The bagging (bootstrap aggregating) is a ‘sampling' algorithm designed to improve the stability and accuracy of algorithms \cite{breiman1996bagging}.
We run the RRCF algorithm with bagging  by choosing 
a \textit{sample size} and 
a \textit{number of iterations}.

\vspace{0.3cm}
\noindent\textbf{Bagging algorithm.}\\
Recall that our data set $X \subseteq \mathbb{R}^d$ is a finite set of $n$ elements.

\noindent\textbf{Step 1.}
Choose the sample size $s \in \{1,2,\dots, n\}$ and the number of iterations $N \in \mathbb{N}$.
Let $k$ be the greatest integer less than or equal to $\frac{n}{s}$, that is, 
$k= \lfloor \frac{n}{s} \rfloor$. Let $j$ be the number of repetitions of \textbf{Step 2}.

\noindent\textbf{Step 2.}
We construct $k$ samples $S_{i,j},\dots, S_{k,j}$ where each $S_{1,j}$ consists of $s$ points, randomly chosen from $X$.
Apply the RRCF algorithm to $S_{1,j},\dots, S_{k,j}$.
For each sample $S_{i,j}$, we obtain an RRCT, $T_{i,j}$ and  
$\textbf{CODISP}(x,S_{i,j},T_{i,j})$ for each $x \in T_{i,j}$.

\noindent\textbf{Step 3.}
Repeat \textbf{Step 2} $N$ times and obtain an RRCF, $F_j$
\[
F_j = \{T_{1,j}, T_{2,j} , \dots ,T_{k,j} \} 
\]
for $j=1,2,\dots, N$.
Then we have the  \textbf{average collusive displacement} of $x\in X$,
\[
\textbf{Avg-CODISP}(x,N)= \frac{1}{a(x)}\sum\limits_{j=1}^N  \sum\limits_{T_{i,j}\in F_j  }  \textbf{CODISP}(x,S_{i,j},T_{i,j})
\]
where
\[
a(x)= \sum\limits_{j=1}^N \sum\limits_{T_{i,j}\in F_j  } \vert  \{x\} \cap S_{i,j} \vert
\]
is the number of times $x$ appears in the bagging algorithm.
In most cases, the RRCF algorithm is employed with bagging. 
%Therefore, one may opt for using $\textbf{CODISP}(x)$ instead of $\textbf{Avg-CODISP}(x)$ shortly.

\subsection{RRCF for time series data}\label{RRCFtime} 
Given an RRCT $T$ of a data set $X$, we can construct a new tree $T'$ by deleting a point from $X$
or inserting a new point to $X$.
In this way, one can apply the RRCF algorithm to time series data effectively.

\subsubsection{Deleting a point from a tree}
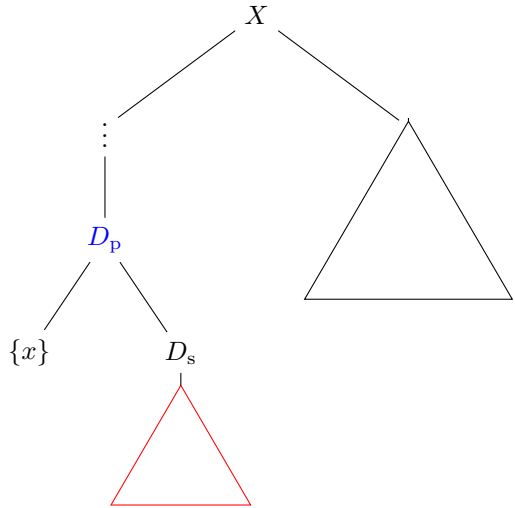
\begin{figure}[h]
\centering

\begin{tikzpicture}
[
level 1/.style = {sibling distance = 4cm},
level 2/.style = {sibling distance = 2cm},
level 3/.style = {sibling distance = 2cm}
]
\node  {$X$}
		child { node { $\vdots$} 
			child { node [blue]{ $D_{\text{p}}$}     
				child { node {$\{x\}$}} 
				child { node {$D_{\text{s}}$}  
					child { node[red, amp1] {}}				
							 }}}
		child {node {}
			child { node[amp3] {}}}  ;
 \end{tikzpicture}
\caption{The graph of $T$}
\label{fig3}
\end{figure}

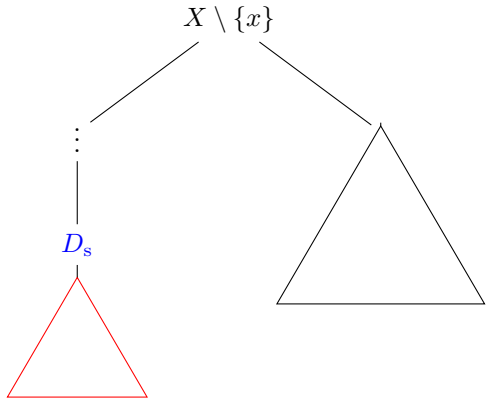
\begin{figure}[h]
\centering

\begin{tikzpicture}
[
level 1/.style = {sibling distance = 4cm},
level 2/.style = {sibling distance = 2cm},
level 3/.style = {sibling distance = 2cm}
]
\node  {$X \setminus \{x\}$}
		child { node { $\vdots$} 
			child { node [blue]{$D_{\text{s}}$}  
					child { node[red, amp1] {}}				
							 }}
		child {node {}
			child { node[amp3] {}}}  ;
 \end{tikzpicture}
\caption{The graph of $T'$}
\label{fig4}
\end{figure}

Let $T$ be an RRCT of a data set $X \subseteq \mathbb{R}^d$ 
and let $x\in X$.
The tree $T'$ obtained by deleting the point $x$ is defined by 
\[
T'=\{ D \setminus \{x\} \mid D\in T  \}.
\]
The graph of $T'$ can be visualized
by deleting the node of $\{ x \}$ in $T$  and by replacing the parents node $D_\text{p}$ of $\{x\}$ with 
the subtree of the sibling node $D_{\text{s}}$ of $D$ (Figs. \ref{fig3} and \ref{fig4}).

\subsubsection{Inserting a point to  a tree}

Let $T$ be an RRCT of a data set $X \subseteq \mathbb{R}^d$.
Let $y \in \mathbb{R}^d \setminus X$ be a new point.

\noindent \textbf{Building process of a tree with a new point.}\\
\noindent \textbf{Step 1.}
Consider the bounding box of $X' =X \cup \{ y \}$
\[
B(X') = \prod\limits_{q=1}^d [\min X_q', \max X_q']
\]
and the total length $l'$ of ‘edges' of the box
\[
l'= \sum \limits_{q=1}^d  ( \max X_q'  - \min X_q' ).
\]

\noindent \textbf{Step 2.}
Let $R$ be a random variable that is uniformly distributed on  $[0,l']$.
Choose $r\in [0,l']$ from the random variable $R$ and obtain the values $s$ and $c$:
\begin{displaymath}
\begin{split}
	  s =& \text{argmin}\left \{t \in \{1,2, \dots, d\} \mid \sum\limits_{i=1}^{t}(\max X_i -\min X_i) \ge r  \right\}, \\
	 c =& \min X_q'+r - \sum\limits_{i=1}^{s-1} (\max X_i' -\min X_i').
\end{split}
\end{displaymath}

\noindent \textbf{Step 3.} 
There are two cases:
\begin{itemize}
\item[\textbf{Case 1}.] $c \notin [\min X_s, \max X_s]$ (Fig. \ref{fig5}). \\
In this case, $X'$ has a fully grown binary tree $T'= \{ X', T, \{y \}\}$ of $X'$.

\begin{figure}[h]
\centering
{
\begin{tikzpicture}
[
level 1/.style = {sibling distance = 3cm},
level 2/.style = {sibling distance = 1cm},
level 3/.style = {sibling distance = 3cm}
]
\node  {$X'$}
		child { node { $\{y\}$}}
		child { node{}
			child { node[amp2] {$T$}}}
 ;
 \end{tikzpicture}
}
\caption{The graph of $T'$ when $c \notin [\min X_s, \max X_s] $ }	
\label{fig5}
\end{figure}
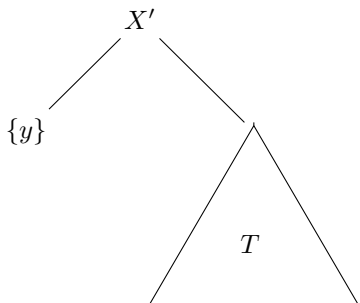
\item[\textbf{Case 2}.] $c \in [\min X_s, \max X_s] $ (Fig. \ref{fig6}). \\
Choose the same dimension cut $s$ and the split value $p$ 
in the partitioning process for $T$ used to partition $X= X' \setminus \{y\}$ and apply the partition process for RRCTs
to $X'$ with the values of $s, p$.
Then we have a binary tree $T'=\{X', A, B  \}$ 
where $y \in A$ and $A \cup B = X'$.
\begin{figure}[h]
\centering
{
\begin{tikzpicture}
[
level 1/.style = {sibling distance = 3cm},
level 2/.style = {sibling distance = 1cm},
level 3/.style = {sibling distance = 3cm}
]
\node  {$X'$}
		child { node { $A$}}
		child { node{$B$}}
 ;
 \end{tikzpicture}
}
\caption{The graph of $T'=\{X', A, B  \}$}
\label{fig6}
\end{figure}
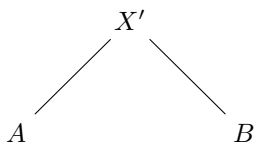
Note that $B$ is also a node in $T$ whose parent is $X$. We update the binary tree $T'$ by replacing the node $B$ with the subtree $T_B$ of $B$ in $T$ (Fig. \ref{fig7}).
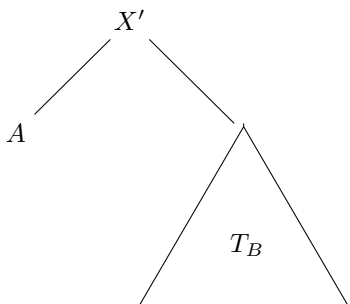
\begin{figure}[h]
\centering
{
\begin{tikzpicture}
[
level 1/.style = {sibling distance = 3cm},
level 2/.style = {sibling distance = 1cm},
level 3/.style = {sibling distance = 3cm}
]
\node  {$X'$}
		child { node { $A$}}
		child { node{}
			child{ node[amp2]{$T_B$}}};
 \end{tikzpicture}
 }
\caption{The graph of   $T'=\{X', A \} \cup T_B$}
\label{fig7}
\end{figure}
\end{itemize}
To build an RRCT $T'$ of $X'$, 
apply this process repeatedly to the output $A$ in  $\textbf{Case 2}$ of $\textbf{Step 3}$
by replacing $X'$ with $A$  in  $\textbf{Step 1}$ and  
by replacing $X$ with $A \setminus \{y\}$ in $\textbf{Step 3}$
until $\{y\}$ becomes an external node.

\begin{remark}
\;
\begin{enumerate}[label=\roman*)]
	\item If \textbf{Case 1} of \textbf{Step 3} holds, $\{y\}$ becomes an external node.
\item In  \textbf{Case 2} of \textbf{Step 3}, the node $A$ containing $y$ has 
less elements than $X'$ and thus 
the process terminates and $\{y\}$ must become an external node in finite time.

\end{enumerate}

\end{remark}

\subsubsection{Application to time series data}
A time series data $X=\{ x_t \}_{t=1}^n \subseteq \mathbb{R}$ is a finite sequence with respect to time $t$. 
To apply the RRCF algorithm to the time series data $X$, 
we use the sliding window technique \cite{CHU1995147} and set
$\text{shingle size} = h,
\text{window size} = w, \text{ and }
\text{forest size} = r$.

The following steps show the detailed procedure of how to apply the RRCF algorithm to time series data.

\noindent\textbf{Step 1.} We make a new data set $S$ which can be obtained from $X$ by constructing $\textit{shingles}$ 
and apply the RRCF algorithm to the new set $S$: 

\noindent Choose a \textbf{shingle size} $1 \le h \le n$ and obtain $n-h+1$
 \textbf{shingles} $s_t \in \mathbb{R}^h$ defined by
\begin{displaymath}
\begin{split}
	  s_{1}&=(x_{1}, x_{2}, \dots, x_{h}),\\
  s_{2}&=(x_{2}, x_{3}, \dots, x_{h+1}),\\
& \;\;\vdots \\
  s_{n-h+1}&=(x_{n-h+1}, x_{n-h+2}, \dots, x_{n}).
\end{split}
\end{displaymath}
 Instead of $X$, we apply the RRCF algorithm with bagging to
\[
S= \{s_t \in \mathbb{R}^h \mid t=1,2,\dots, n-h+1 \}.
 \]

\noindent\textbf{Step 2.}
Choose a \textbf{window size}  $w$  and a forest size $r$.
Let $n'=n-h+1$ be the number of elements in $S$.
For $i=1,2, \dots, n'-w+1$, we have a \textbf{window}
\[W_i=\{s_i, s_{i+1}\dots, s_{i+w-1} \} \subseteq S.\]
Construct an RRCF $F_1$ which has $r$ RRCTs of $W_1$.
For each RRCT $T \in F_i$ with $i \ge 1$, we obtain a ‘next' RRCT $T' \in F_{i+1}$ 
in the following way:
\begin{enumerate}[label = \roman*)]
\item Delete a point $s_i$ from $T \in F_i$ and get the tree $T''$.
\item  Insert a point $s_{i+w}$ to the tree $T''$ and get a tree $T'$ of $W_{i+1}$.
\end{enumerate}
Then we have the \textbf{collusive displacement} of $s_j$ \cite{guha2016robust} given by
\[
\begin{cases}
\frac{1}{j}
\sum\limits_{i=1}^{j}\left( \frac{1}{r}\sum\limits_{T \in F_i}
\textbf{CODISP}(s_j,W_i,T) \right) & \text{ if $1\le j \le w$},\\
 \frac{1}{w}
\sum\limits_{i=j-w+1}^{j}\left( \frac{1}{r}\sum\limits_{T \in F_i}
\textbf{CODISP}(s_j,W_i,T) \right) & \text{ if $w< j \le n'-w$},\\
 \frac{1}{n'-j+1}
\sum\limits_{i=j-w+1}^{n'-w+1}\left( \frac{1}{r}\sum\limits_{T \in F_i}
\textbf{CODISP}(s_j,W_i,T) \right) & \text{ if $n'-w< j \le n'$}.
\end{cases}
\]
When we apply the RRCF algorithm to time series data, we  will  write $\textbf{CODISP}(x)$ 
denoting $\textbf{Avg-CODISP}(x)$.

\begin{remark}
\;
\begin{enumerate}[label=\roman*)]
\item To obtain the \textbf{Avg-CODISP} of $s_j$, 
we calculate every \textbf{CODISP} of $s_j$ over all windows which contain $s_j$.
\item Suppose we have a new point $x_{n+1}$.
 Put 
$s_{n'+1}= s_{n-m}=(x_{n-m},\dots, x_{n+1})$.
Then we have the new window
\[W_{n'-w+2}=\{s_{n'-w+2}, s_{n'-w+3}, \dots, s_{n'+1} \} \subseteq S \cup \{s_{n'+1} \}.\]
We construct an RRCF $F_{n'+1}$ in the same way and have
\[
\textbf{Avg-CODISP}(s_{n'+1})=
\frac{1}{r}\sum\limits_{T \in F_{n'-w+2}}
\textbf{CODISP}(s_j,W_{n'-w+2},T).
\]

\end{enumerate}
	
\end{remark}

\section{Density Measure}\label{density measure}
To introduce our proposed WIF (weightd IF) and WRCF (weighted RCF) in Section \ref{Weighted Isolation Forest and Weighted Random Cut Forest}, we begin by introducing the density measure. We first address the motivation behind the density measure through an example, illustrating certain issues with ordinary partitioning (referenced in Section  \ref{motivation}). Following that, we formally define the density measure in Section \ref{Definitions}. In Section \ref{propositions}, we will demonstrate that the density measure we define is  invariant under scaling and translation and possesses the property that its values approach 1 when the data is  denser than a uniform distribution and approach 0 when the data is  sparser than or close to a uniform distribution.

\subsection{Motivation}\label{motivation}
For an isolation tree or an RRCT from a given data set $X$, 
we have used the split value $p$, obtained from the random variable
\[
P|_{Q=q} \sim \textit{Uniform}[m , M]
\]
where $q$ is  a dimension cut and $m=\min X_q$, $M=\max X_q$.
Note that the random variable $P|_{Q=q}$ is determined by only two values $m$ and $M$. 
Both the IF and RRCF algorithms do not take the given data structure into account when choosing the split values.

\begin{table}
\centering
\begin{tabular}{ |c|c|c| } 
\hline
& IF & RRCF \\
\hline
Dimension Cuts & $\text{Prob}^{\text{IF}}(Q=q) = \frac{1}{d}$ & $\text{Prob}^{\text{RRCF}}(Q=q) = \frac{l_q}{l}$  \\
\hline
Split Values &  \multicolumn{2}{|c|}{randomly chosen in $[m, M]$} \\
\hline
\end{tabular}
\caption{The method for determining dimension cuts and split values in the IF and RRCF algorithms}	
\label{IFRRCF}
\end{table}

In Table \ref{IFRRCF}, we see that the RRCF uses improved dimension cuts but it still uses  the same method as the IF algorithm when 
determining the split values.
Example \ref{ex1} illustrates this particular issue.

\begin{example} \label{ex1}
Consider a set $X \subseteq \mathbb{R}^2$ of 6 points given by
\[ 
X= \{ (0,0), (1,0), (2,0), (10,0), (11,0), (12,0)\}.
\]
Note that the $y$-coordinate values of $X$ are  all zero, that is, $\min X_2 =\max X_2 =0$.
Therefore, $\text{Prob}(Q=2)=0$ and so we only  consider $P|_{Q=1}$. 
Considering the ‘shape' of $X$, 
one might think there are two clusters 
\begin{displaymath}
\begin{split}
	  S_1=& \{ (0,0), (1,0), (2,0)\},  \\
	  S_2=& \{ (10,0), (11,0), (12,0)\}
\end{split}
\end{displaymath}
and want to have
the binary tree $T= \{X ,S_1, S_2\}$ as in Fig. \ref{fig8}.

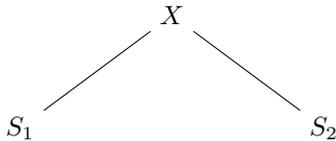
\begin{figure}[h]
	 \begin{center}
\begin{tikzpicture}
\node  {$X$}
[
level 1/.style = {sibling distance = 4cm},
level 2/.style = {sibling distance = 2cm}
]
		child {node {$S_1$}}
		child{node {$S_2$}}
	;
\end{tikzpicture}
\end{center}
\caption{The desired graph of $T$ (Example \ref{ex1})}\label{fig8}
\end{figure}
Having the tree $T$ in Fig. \ref{fig8} is equivalent to using the split value $p \in (2, 10)$. But if we choose the split value $p$ obtained from the random variable $P|_{Q=1}$, which is  uniformly distributed on $[0,12]$, we have 
\[
	  \text{Prob}(2 <P|_{Q=1}< 10 )= \frac{8}{12}=\frac{2}{3}
\]
which is the probability of building the tree $T$ in Fig. \ref{fig8} at the first partitioning process.
In other words, we would miss the tree with the probability  of $\frac{1}{3}$.
\end{example}

\subsection{Definitions}\label{Definitions}
As seen in \text{Example \ref{ex1}}, the ordinary partitioning algorithm does not consider the shape of $X$. To address this limitation and take into account the shape of the data, we introduce a new partitioning algorithm called \textit{Density-Aware Partitioning} (Section \ref{Density-Aware Partitioning}). This algorithm is designed to overcome the mentioned issue. Before introducing the new method, we define a density measure $\mu$ that quantifies the ‘density' of the given data set $X$.

\begin{definition}
Let $F_0$ be the collection of all nonempty finite sets in $\mathbb{R}$.
For each
$Y \in F_0$,
we define the \text{radius} of $Y$ by
\[
	  \varepsilon =\begin{cases}
 \frac{\max Y-\min Y}{2(\vert Y 
\vert-1)} & \text{if $\vert Y \vert \neq1$,}\\
0 & \text{if $\vert Y \vert =1$.}
\end{cases}
\]
 Denote  the half-open interval of radius $\varepsilon$ centered at $p$ by
\[
I_{p,\varepsilon} =
\begin{cases}
[ p-\varepsilon, p+\varepsilon)  &\text{ if $\varepsilon>0$},	\\
\{p\}  &\text{ if $\varepsilon=0$}.
\end{cases}
\]
The \textbf{density measure} $\mu_0$ on $\mathbb{R}$ is a map defined by
\begin{displaymath}
\begin{split}
	  \mu_0:   F_0 &\rightarrow [0,1] \\ 
        				Y & \mapsto \max \left\{ \frac{\vert  I_{p,\varepsilon} \cap Y \vert }{\vert Y \vert}\mid  p \in [\min Y,  \max Y ] \right\}.
\end{split}
\end{displaymath}

\end{definition}

\begin{definition}

Let $F$ be the collection of all nonempty finite sets in $\mathbb{R}^d$. 
The \textbf{density measure} $\mu$ on $\mathbb{R}^d$ is  a map defined by
\begin{displaymath}
\begin{split}
	  \mu:   F &\rightarrow [0,1] \\
        				X & \mapsto \frac{1}{d} \sum\limits_{q=1}^d \mu_0(X_q).
\end{split}
\end{displaymath}
The definition of $X_q$ is provided in Section \ref{IsolationForest}, specifically in Eq.  (\ref{eq1}).
\end{definition}

\begin{remark}\label{remark5}
\;\\
If $Y=\{a,2a,\dots,na\}$, which is equally distributed with distance $a>0$, then 
\[
\varepsilon = \frac{an-a}{2(n-1)}=\frac{a}{2}.
\]
Thus the radius of $Y$ is the half of the equal distance $a$
and  $[p-\frac{1}{2}a, p+\frac{1}{2}a) \cap Y$ has only one element for all $p \in [a,na]$. Thus we have 
\begin{displaymath}
\begin{split}
 \mu (Y) 
&=
 \max 
\left\{ \frac{ \vert  [p-\frac{1}{2}a, p+\frac{1}{2}a)  \cap Y \vert}{n} \mid  p \in [a,  na ] \right\}\\
&=
\frac{1}{n}.	  
\end{split}
\end{displaymath}

\end{remark}

Example \ref{ex2} illustrates the density measure for the 2D-defined data with some examples.
\begin{example}\label{ex2} 
Let $C_0$ and $C_2$ be sets of $50$ and $5$ points in $\mathbb{R}^2$ respectively, where 
each point of both sets is obtained from the random variable $(A, B)$ 
such that both $A$ and $B$ are normally distributed.
For each $k>0$, define
\[
C_{1} = \left\{\frac{1}{k}(x,y) \mid (x,y) \in C_0\right \}.
\]
Put $X= C_{1} \cup C_2$. 
For each $k=1,5,30$ and $50$, the densities of $X$ are depicted in Fig. \ref{fig9}. As shown in the figure, an increase in sparsity results in a decrease in the density measure. Note that $X$ with $k=5,30$ and $50$ have all the same bounding box and so the IF and RRCF would treat them \textcolor{black}{equally} when choosing the split values at the first partitioning step.
\begin{figure}[h]
\centering
\begin{subfigure}{0.45\textwidth}
    \centering
    \includegraphics[scale=0.32]{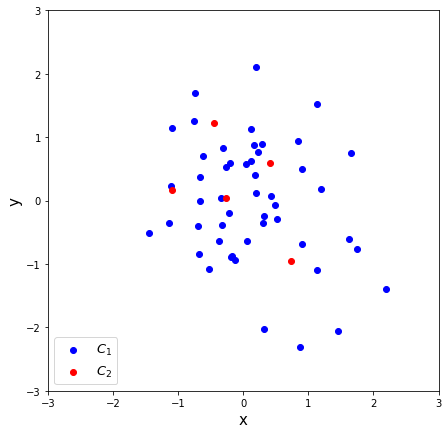}
    \caption{$k=1$, \; $\mu(X)=0.07$}
\end{subfigure}
\begin{subfigure}{0.45\textwidth}
    \centering
    \includegraphics[scale=0.32]{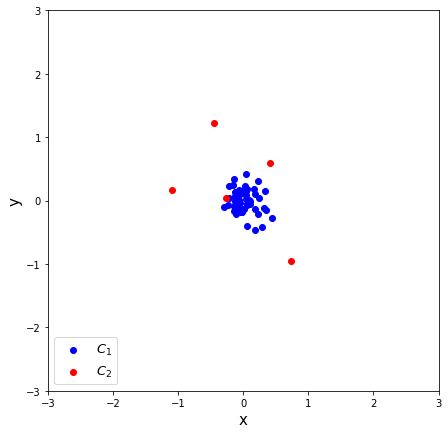}
    \caption{$k=5$,  \; $\mu(X)=0.35$}
\end{subfigure}
\begin{subfigure}{0.45\textwidth}
    \centering
    \includegraphics[scale=0.32]{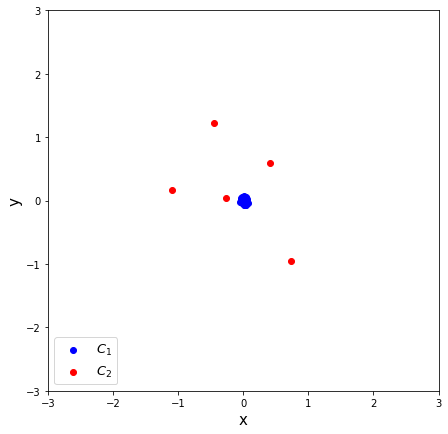}
    \caption{$k=30$, \; $\mu(X)=0.48$}
\end{subfigure}
\begin{subfigure}{0.45\textwidth}
    \centering
    \includegraphics[scale=0.32]{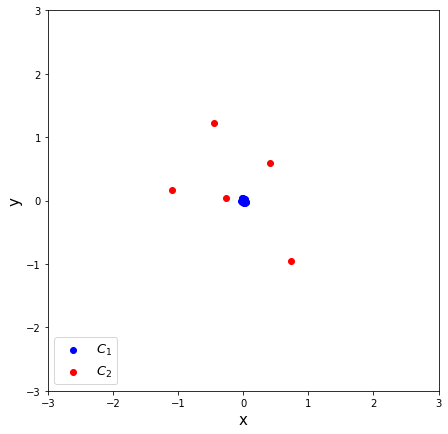}
    \caption{$k=50$,\;  $\mu(X)=0.7$}
\end{subfigure}
\caption{The graph of $X=C_1 \cup C_2$ (Example \ref{ex2})}
\label{fig9}
\end{figure}
\end{example}

\subsection{Propositions}\label{propositions}
We investigate the properties of the proposed density measure. Propositions \ref{prob2} and \ref{prop3} examine two extreme cases of $X_q$ with respect to the density measure $\mu$. Propositions \ref{prop4} and \ref{prop5} demonstrate that the density measure is invariant under reflection, scaling, and translation.
\begin{proposition}\label{prob2}
Fix $q \in \{1,2,\dots, d\}$. The following statements are equivalent:
\begin{enumerate}[label=\roman*)]
\item $X_q$ is uniformly distributed, that is,
$ X_q  =\{ak+b \mid k=1,2,\dots, n \}$ for some $a>0, b \in \mathbb{R}$.
\item $\mu(X_q)=\frac{1}{n}$  where $n$ is the number of elements in $X_q$.
\end{enumerate}
\end{proposition}

\begin{proof} Let $m=\min X_q$ and $M= \max X_q$.

\begin{enumerate}[label=\arabic*)]
\item	i) $\Rightarrow$ ii): Suppose that i) holds. Then we have
\[
\varepsilon_q = \frac{M - m}{2(n-1)} = \frac{a(n-1)}{2(n-1)}=\frac{a}{2}
\]
and therefore 
\[
 I_{p,\varepsilon_q} =\left[p -\frac{a}{2},  p +\frac{a}{2} \right).
\]
If we write $x_k=ak+b$ for $k=1,2,\dots, n$, then
\[
x_{k+1} - x_k = a
\]	
for $k \in \{1,2,\dots, n-1\}$.
For  any $p \in [x_k, x_{k+1} )$ with $k \in \{1,2,\dots, n-1\}$, 
\begin{displaymath}
\begin{split}
 I_{p,\varepsilon_q} \cap X_q =  \left[p -\frac{a}{2},  p +\frac{a}{2} \right) \cap  X_q= \{x_k\}
\end{split}
\end{displaymath}
which shows 
$ \vert I_{p,\varepsilon_q}  \cap X_q\vert =1$
for all  $  p \in [m,  M ]$.
Thus we have
\[
\mu(X_q)=
\max 
\left\{ \frac{ \vert  I_{p,\varepsilon_{q}} \cap X_q \vert}{n} \mid  p \in [m,  M ] \right\} 
=\frac{1}{n}.
\] 
\item
ii) $\Rightarrow$ i): Suppose $\mu (X_q)= \frac{1}{n}$. 
Then $\vert  I_{p,\varepsilon_{q}} \cap X_q \vert = 1$ for all  $  p \in [m,  M ]$.
Write $X=\{x_1,\dots,x_n\}$ with $x_1 \le x_2 \le \dots \le x_n$.
Then we must have
\[
x_k = x_1+ \varepsilon_q(k-1)
\]
for $k=1,2,\dots, n-1$. 
This proves i) by taking $a=\varepsilon_q>0$ and $b=x_1-\varepsilon_q$.
\end{enumerate}
\end{proof}
Note that Proposition \ref{prob2} is the  generalization of Remark \ref{remark5}.

\begin{proposition}\label{prop3}
Fix $q \in \{1,2,\dots, d\}$.
The following statements are equivalent:
\begin{enumerate}[label=\roman*)]
\item $X_q$ is a multiset where all the elements are identical.
\item $\mu(X_q)=1$.
\end{enumerate}
\end{proposition}
\begin{proof}
 Let $m=\min X_q$ and $M= \max X_q$.
\begin{enumerate}[label=\arabic*)]
\item	i) $\Rightarrow$ ii): Suppose that $X_q=\{x,\dots,x\}$ with $n$ duplicates. 
Since $m = x = M$,  we have $\varepsilon_{q} = 0$ and  $I_{p,0}=\{x\}=X_q $. 
Then
\begin{displaymath}
\begin{split}
	\mu(X)
=& \frac{1}{d}\sum_{q=1 }^d \max 
\left\{ \frac{ \vert  I_{p,\varepsilon_{q}} \cap X_q \vert}{n} \mid  p \in [m,  M ] \right\} \\
=& \frac{1}{d}\sum_{q=1 }^d \max 
\left\{ \frac{ \vert  X_q \cap X_q \vert}{n}  \right\} \\
=& \frac{1}{d}\sum_{q=1 }^d \max 
\left\{ \frac{ n}{n}  \right\} \\
=& 1.
\end{split}
\end{displaymath}

\item
ii) $\Rightarrow$ i): Suppose $\mu (X_q)=1$.  
Clearly, i) holds for $n=1$. Assume that $n \ge 2$. 
Since $\mu (X_q)=1$, we must have
\[ \max 
\left\{ \frac{ \vert  I_{p,\varepsilon_{q}} \cap X_q \vert}{n} \mid  p \in [m,  M ] \right\} =1.
\]
Then there exist $p_0 \in  [m,  M ] $ such that 
$\vert  I_{p_0,\varepsilon_{q}} \cap X_q \vert =n$, that is,
$X_q \subseteq  I_{p_0,\varepsilon_{q}} $.
Suppose $m<M$.  Then
 \[
 I_{p_0,\varepsilon_{q}} = \left [p_0- \frac{M - m}{2(n-1)} , p_0 +\frac{M -m}{2(n-1)} \right)
\] must contain both $m$ and $M$ and thus 
\[
p_0- \frac{M- m}{2(n-1)}\le m< M< p_0 +\frac{M -m}{2(n-1)}.
\]
Then we have
\[
\frac{(2n-3)M +m}{2(n-1)} < p_0 \le \frac{M+(2n-3)m}{2(n-1)}.
\]
By comparing the left and right sides of the above inequality, we have
\begin{displaymath}
\begin{split}
	  &\frac{(2n-3)M +m}{2(n-1)} < \frac{M+(2n-3)m}{2(n-1)} \\
\; \Leftrightarrow \;  	  &(2n-3)M +m < M+(2n-3)m\\
\; \Leftrightarrow \; &(2n-4)M<(2n-4)m. \\
\; \Leftrightarrow \; &M \le m. \\
\end{split}
\end{displaymath}
Hence, we must have \(m = M\), that is, all the elements of $X_q$ are identical. Therefore, we have $X_q=\{p_0,\dots,p_0\}$ with $n$ duplicates. 

\end{enumerate}
\end{proof}

\begin{proposition}\label{prop4}
	The density measure $\mu$ is invariant with respect to reflection, that is,
\[
\mu(X)=\mu(-X).
\]
\end{proposition}

\begin{proof}
Fix $q\in \{1,2,\dots,d\}$. 
Clearly, we have
\begin{displaymath}
\begin{split}
\varepsilon_{q,X}
=\frac{-\min X_q - (-\max X_q ) }{2n-1}
=\varepsilon_{q,(-X)}.
\end{split}
\end{displaymath}
Let $p_0\in [\min X_q , \max X_q] $ such that 
$\mu_0(X_q)=\frac{ \vert  I_{p_0,\varepsilon_{q,X}} \cap X_q \vert}{\vert X_q \vert}$.
Write $ I_{p_0,\varepsilon_{q,X_q}} \cap X_q=\{ a_1,a_2,\dots,a_k\}$. Then we have
\begin{displaymath}
\begin{split}
	  \mu_0 (X_q)
&=\frac{ \vert \{ a_1,a_2,\dots,a_k \} \vert}{\vert X_q \vert} \\
&=\frac{ \vert  \{-a_1,-a_2,\dots, -a_k \} \vert}{\vert (-X)_q \vert} \\
	&=  \mu_0 ((-X)_q)
\end{split}
\end{displaymath}
where $q \in \{1,\dots,d\}$ can be arbitrary. Hence, $\mu(X)=\mu(-X)$.
\end{proof}

\begin{proposition}\label{prop5}
The density measure $\mu$ is invariant with respect to scaling and translation.
  That is, 
\[
\mu(X) = \mu(aX+b)
\]	
for any $a \neq 0, b \in \mathbb{R}$. 
\end{proposition}

\begin{proof}
Fix $q\in \{1,2,\dots, d\}$.
If $X_q$ has $n$ duplicates,  then  $aX_q+b$ also has $n$ duplicates.
By \textbf{Proposition \ref{prop3}},  we have
\[
\mu_0(X_q) = 1 = \mu_0(aX_q+b).
\]
By \textbf{Proposition \ref{prop4}}, $\mu_0(aX_q)=\mu_0(-aX_q)$ and so
we may assume that $a>0$. Suppose that $\min X_q < \max X_q$ and let
\[ Y_q=aX_q+b = \{ax+b \mid x \in X_q \}.\]
 Then we have
\begin{displaymath}
\begin{split}
	\varepsilon_{q,Y}
&=\frac{(a \max X_q +b) - (a \min X_q +b)}{2(n-1)}\\
&=a\varepsilon_{q,X}  
\end{split}
\end{displaymath}

and 
\begin{displaymath}
\begin{split}
&   x \in I_{p,\varepsilon_{p,X}} \cap X_q  \\
 \Leftrightarrow \;
 	&x \in  [ p-\varepsilon_{q,X}, p+\varepsilon_{q,X} ) \cap X_q \; \\
 \Leftrightarrow \;& ax +b  \in  [ a(p-\varepsilon_{q,X})+b, a(p+\varepsilon_{q,X})+b ) \cap (aX_q+b) \\
 \Leftrightarrow \; & y \in  [ p' - \varepsilon_{q,Y}, p' + \varepsilon_{q,Y} ) \cap Y_q 
\text{ with $p'=ap+b$} \\
 \Leftrightarrow \; & y \in  I_{p',\varepsilon_{q,Y}} \cap Y_q.
\end{split}
\end{displaymath}
Therefore,
\begin{displaymath}
\begin{split}
	\mu_0(X_q) &= \max 
\left\{ \frac{ \vert  I_{p,\varepsilon_{q,X}} \cap X_q \vert}{n} \mid  p \in [\min X_q, \max X_q ] \right\} \\
& = \max 
\left\{ \frac{ \vert  I_{p',\varepsilon_{q,Y}} \cap Y_q \vert}{n} \mid p'\in [\min Y_q,  \max Y_q ] \right\} \\
&=\mu_0(Y_q)
\end{split}
\end{displaymath}
where $q \in \{1,\dots,d\}$ can be arbitrary. Hence, $\mu(X)=\mu(aX+b)$.
\end{proof}

\section{Weighted Isolation Forest and Weighted Random Cut Forest} \label{Weighted Isolation Forest and Weighted Random Cut Forest}
A \textit{Weighted Isolation Forest} (WIF) and a \textit{Weighted Random Cut Forest} (WRCF) are similar to the IF and RRCF, respectively. The key distinction lies in how the new algorithms, WIF and WRCF, select the split value $p$ through \textit{density-aware partitioning}. The goal of density-aware partitioning is to avoid ‘bad' split values located in a cluster in the data set.

\subsection{Density-Aware Partitioning}\label{Density-Aware Partitioning}
Suppose that the dimension cut $q \in \{1,2,\dots, d\}$ is chosen.
Choose an integer $\alpha \ge 2$.

\noindent\textbf{Stage 1.}
Choose the split value $p$ as usual from the random variable 
\[P|_{Q=q} \sim \textit{Uniform}[\min X_q,  \max X_q].\]

\noindent\textbf{Stage 2.}
If $\vert I_{p,\varepsilon_q} \cap X_q \vert \ge \alpha $, we repeat $\textbf{Stage 1}$ 
until we choose $p$ such that
\[\vert I_{p,\varepsilon_q} \cap X_q \vert   < \alpha .\]

\begin{remark}
\;
\begin{enumerate}[label=\roman*)]
\item A split value $p$ with $\vert I_{p,\varepsilon_q} \cap X_q \vert \ge \alpha $ 
is regarded as being in a cluster.
\item 	
 The probability that
 the partitioning process does not terminate within $n$ trials is
\[
(\text{Prob}(\vert I_{P|_{Q=q},\varepsilon_q} \cap X_q \vert \ge \alpha ))^n.
\]
We will show later in \textbf{Theorem }\ref{thm1} that this probability is highly small 
and we do not have to worry about the infinite loop in \textbf{Stage 2}.
\end{enumerate}

\end{remark}

Recall \text{Example} \ref{ex1} in \textbf{Section} \ref{motivation}.
In this case, $X$ has high density with $\mu(X)=0.75$.
We will build a binary tree with the new process of choosing the split value $p$. 
Note that
\[ 
\varepsilon_{1}=\frac{12-0}{2(6-1)}=1.2.
\]
Define a map $f$ as 
\begin{displaymath}
\begin{split}
	  f: [0,12] & \rightarrow \mathbb{Z}  \\
p & \mapsto \vert I_{p, \varepsilon_1} \cap X_1 \vert.
\end{split}
\end{displaymath}

\begin{figure}[h]
\centering
\includegraphics[scale=0.4]{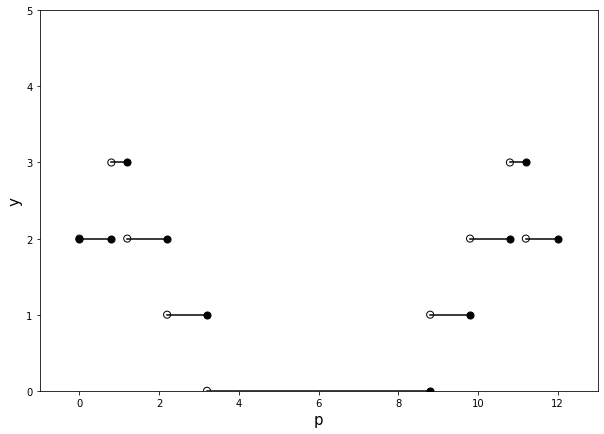}
\caption{The graph of $y=f(p)$}
\label{fig10}
\end{figure}

Then we have
\begin{displaymath}
\begin{split}
	 f(p)= \vert I_{p, \varepsilon_1} \cap X_1 \vert = 
\begin{cases}
2 & \text{ if $0 \le p\le 0.8$} \\	
3 & \text{ if $0.8 < p \le 1.2$} \\	
2 & \text{ if $1.2 < p \le 2.2$} \\	
1 & \text{ if $2.2 < p \le 3.2$}  \\
0 & \text{ if $3.2 \le p < 8.8$}  \\
1 & \text{ if $8.8 \le  p \le 9.8$}  \\
2 & \text{ if $9.8 < p\le 10.8$}  \\
3 & \text{ if $10.8 < p \le 11.2$}  \\
2 & \text{ if $11.2 < p \le 12$}  \\
\end{cases}
\end{split}
\end{displaymath}
with the graph shown in Fig. \ref{fig10}.

\begin{figure}[h]
\centering
			       \includegraphics[scale=0.4]{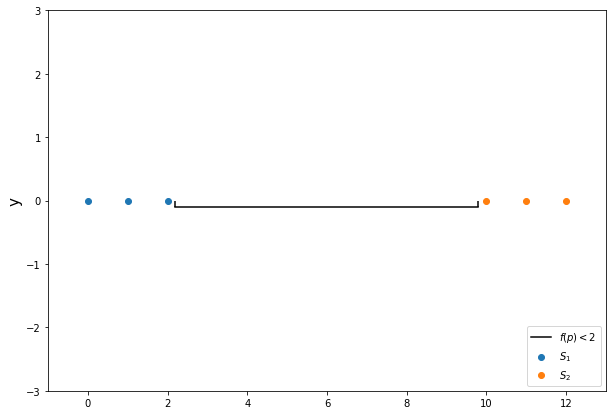}
\caption{The graph of points in $X$ (Example \ref{ex1})}
\label{fig11}
\end{figure}

\noindent 
If we take $\alpha >3$,  we always have
$
\vert I_{p, \varepsilon_1} \cap X_1 \vert < \alpha
$
 and thus 
we never repeat \textbf{Stage 1}. Suppose $\alpha =2$ and note that
\[
	f(p)=\vert I_{p, \varepsilon_1} \cap X_1 \vert <2  \; \Leftrightarrow \; 2.2 < p \le 9.8.
\]
Figure \ref{fig11} illustrates the case with the six points from \text{Example \ref{ex1}}. The extended solid line represents the interval where $f(p)<2$.
\noindent We \textit{always} have the binary tree $T= \{X ,S_1, S_2\}$ (Fig. \ref{fig8}) with the density-aware partitioning ($\alpha =2$).
The probability that we repeat \textbf{Stage 1}  is 
\[ 
\frac{\mu_{\text{Bor}}(A)}{\mu_{\text{Bor}}([0, 12])}=\frac{(2.2-0)+ (12-9.8)}{12-0}=\frac{11}{30}\]
where $\mu_{\text{Bor}}$ is the Borel measure  \cite{folland2013real} on $\mathbb{R}$ and
$
A=\{ p \in [0, 12]: f(p) \ge 2 \}$.
Hence, the probability that we repeat \textbf{Stage 1} $N$ times  is 
$\left ( \frac{11}{30}\right)^N$.
In other words, the probability that the new process terminates within $N-1$ trials is
$1-(\frac{11}{30})^N$. In  Table \ref{table2}, we find that the new process terminates within 3 trials with the probability of $98.2 \%$. In Section \ref{Proposition and Theorem}, we will show that  this probability has a lower bound (Table \ref{table3}).
\begin{table}
\centering
\begin{tabular}{|c|c|c|c|c|c|c|c|c|}
\hline
$N$& 1& 2& 3 &  4 & 5 \\
\hline
$1- \left(\frac{11}{30}\right)^N$& 0.63& 0.87 & 0.95&  0.98 & 0.99 \\
\hline	
\end{tabular}
\caption{The probability that the new process terminates within $N-1$ trials }
\label{table2}
\end{table}

\subsection{WIF and WRCF}

Recall that the IF and RRCF use an ordinary partitioning process (Algorithm \ref{Ordinary Partitioning_algorithm}) where split values are determined without considering the shape or density of the data. The WIF and WRCF are obtained by choosing split values with density-aware partitioning (Algorithm \ref{Density-Aware Partitioning_algorithm}). This enhancement allows our proposed algorithms to take into account the shape and density of the data set.

\begin{algorithm}
\caption{Ordinary Partitioning}
\begin{algorithmic}[1]
\State Suppose dimension cut $q \in \{1,2,\dots, d\}$ is chosen.
\State Choose split value $p$ from $P|_{Q=q} \sim \textit{Uniform}[\min X_q, \max X_q]$.
\end{algorithmic}\label{Ordinary Partitioning_algorithm}
\end{algorithm}

\begin{algorithm}
\caption{Density-Aware Partitioning}
\begin{algorithmic}[1]
\State Suppose dimension cut $q \in \{1,2,\dots, d\}$ is chosen.
\State Choose an integer $\alpha \ge 2$.
\Procedure{Stage 1}{}
    \State Choose split value $p$ from $P|_{Q=q} \sim \textit{Uniform}[\min X_q, \max X_q]$.
\EndProcedure
\Procedure{Stage 2}{}
    \While{$\vert I_{p,\varepsilon_q} \cap X_q \vert \ge \alpha$}
        \State \textbf{do} Stage 1
    \EndWhile
\EndProcedure
\end{algorithmic}\label{Density-Aware Partitioning_algorithm}
\end{algorithm}

\subsection{Proposition and Theorem}\label{Proposition and Theorem}
Proposition \ref{pro6} demonstrates the existence of the split value $p$, while Theorem \ref{thm1} establishes that the density-aware partitioning eventually terminates.
\begin{proposition}\label{pro6}
For any $q \in \{1,2,\dots,n\}$ and $\alpha  \ge 2$, there exists $p \in  [\min X_q, \max X_q]$ such that 
$\vert I_{p,\varepsilon_q} \cap X_q\vert < \alpha $.
\end{proposition}

\begin{proof}
We prove this by contradiction. 
Suppose that there exists $q \in \{1,2,\dots,d\}$ such that 
\[
\vert I_{p,\varepsilon_q} \cap X_q\vert \ge \alpha 
\]	
for all $p \in [\min X_q, \max X_q]$. 
 Let $m=\min X_q$ and $M= \max X_q$.

For $k=1,2,\dots, n $, take
\[
p_k=m +2(k-1)\varepsilon_q.
\]
Then  we have
\[
\bigcap\limits_{k=1}^{n}I_{p_k,\varepsilon_{q}} 
=\bigcap\limits_{k=1}^{n}
 [m +(2k-3) \varepsilon_q  , m +(2k-1)\varepsilon_q)
=
 \varnothing
\]
 and
\begin{displaymath}
\begin{split}
\bigcup\limits_{k=1}^{n}I_{p_k,\varepsilon_{q}} 
=&\bigcup\limits_{k=1}^{n}
 [m +(2k-3) \varepsilon_q  , m +(2k-1)\varepsilon_q)\\
=&[m - \varepsilon_q , M + \varepsilon_q) \\
\supseteq & [m, M].	  
\end{split}
\end{displaymath}
But 
$\vert  I_{(p_k,\varepsilon_{q})} \cap X_q \vert \ge \alpha 
$
for all $k=1,2,\dots, n$ and so $X_q$ has at least $\alpha n$ elements.
This is a contradiction since 
$X$ has $n $ elements.
\end{proof}

\begin{theorem}
Fix $q \in \{1,2,\dots,d\}$ and $\alpha  \ge 2$ and let 
\[ A=\{p \in [m ,M] : \vert I_{(p,\varepsilon_q)} \cap X_q \vert \ge \alpha  \}
\]
where  $m=\min X_q$ and $M= \max X_q$ such that $m < M$.
Then we have 
\[
\mu_{\text{Bor}}(A) <  \frac{l_q}{\alpha} 
\] 
where $\mu_{\text{Bor}}$ is the Borel measure on $\mathbb{R}$.	

\label{thm1}
\end{theorem}

%%%%%
\begin{proof}
If $A$ is empty, it is trivial. Otherwise, suppose that $A$ is nonempty.
Consider a map 
\begin{displaymath}
\begin{split}
	  f : \mathbb{R} & \rightarrow \mathbb{Z} \\
	p & \mapsto \vert I_{p,\varepsilon_q} \cap X_q \vert.
\end{split}
\end{displaymath}
Let $A'=f^{-1}([\alpha ,\infty))$ 
so that $A =A'\cap [m,M]$.  
For $p\in A'$, put
\begin{displaymath}
\begin{split}
	  m_p &= \min (I_{p,\varepsilon_q}\cap X_q), \\
M_p &= \max (I_{p,\varepsilon_q} \cap X_q ).
\end{split}
\end{displaymath}
Since $X_q$ is finite,  
\[
	  R =\max \{ M_p - m_p   :     p \in A' \} < 2 \varepsilon_q \]
is well-defined.  
Now let
     \[r = \min \{ \vert x - y \vert \mid x \neq y, \; x,y \in X_q\}. \]
If $r=0$, then $X_q$ has $n$ duplicates and thus $m=M$.
Since $m<M$,  this is impossible. Thus, we have $r >0$.

Since $A \neq \varnothing$, there exists $p_1 \in A'$. 
Then
we have the half-open interval containing $p_1$:
\[B_1 =(a_1, b_1]=(M_{p_1} -\varepsilon_q,m_{p_1} + \varepsilon_q]  \subseteq A'.\]
If $B_1 \cup \dots \cup B_{k-1} \subsetneq A'$ with $k\ge 2$,
there exists 
\[p_k \in \mathbb{R}\setminus (B_1 \cup \dots \cup B_{k-1})  \;\; \text{ such that } \;\;
 \vert I_{(p_k,\varepsilon_q)} \cap X_q \vert \ge \alpha. \]
Similarly, we have  the half-open interval containing $p_k$:
\[B_{k}=(a_k, b_k]=(M_{p_k} -\varepsilon_q,m_{p_k} + \varepsilon_q] \subseteq A'.\]
Since $B_{k } \nsubseteq B_{k-1}$, we have
\[
B_k \setminus B_{k-1} =
\begin{cases}
(a_k, b_k] &  \text{ if $B_k \cap B_{k-1} = \varnothing$},\\
(a_k,  a_{k-1} ]  & \text{ if $a_{k-1} < b_k<  b_{k-1}$},	\\
(b_{k-1},  b_k] & \text{ if  $a_{k-1} <a_{k} < b_{k-1}$ }.
\end{cases}
\]
Since $a_i= M_{p_i} -\varepsilon_q, \; b_i =m_{p_i} + \varepsilon_q$, we have
\begin{displaymath}
\begin{split}
	  & \mu_{\text{Bor}}(B_k \setminus B_{k-1}) \\
\ge & \min ( 2\varepsilon_q -(M_{p_k} - m_{p_k}) , \vert M_{p_{k-1}} - M_{p_k} \vert,  \vert m_{p_k}- m_{p_{k-1}}  \vert) \\
\ge & \min ( 2\varepsilon_q -R, r ). \\
\end{split}
\end{displaymath}
Therefore,
\begin{displaymath}
\begin{split}
	   &\mu_{\text{Bor}}\left(B_1 \cup \dots \cup B_k\right) -\mu_{\text{Bor}}(B_1 \cup \dots \cup B_{k-1}) \\
=& \mu_{\text{Bor}}\left(B_k \setminus B_{k-1}\right) \\
\ge & \min ( 2\varepsilon_q -R, r ) \\
  >& 0
\end{split}
\end{displaymath}
where $\min ( 2\varepsilon_q -R, r ) $ is the fixed positive number for all $k$.
Since $m(A') \le l_q + 2 \varepsilon_q < \infty$, the following sequence 
\[
B_1 \subsetneq (B_1 \cup B_2) \subsetneq \dots \subsetneq (B_1 \cup \dots \cup B_k) 
\subsetneq \dots \subseteq A'
\]
must terminate, that is, $B_1 \cup \dots \cup B_{t}=A'$ for some $t$.
In particular, 
$A'$ is Borel measurable.
Note that a union of half-open intervals is also a union of half-open intervals. 
Then we may assume
$B_i$ are mutually disjoint half open intervals 
of the form $(a_i, b_i]$.
For each $B_i$, take small $r_i>0$ and  put
 $c_i=\left \lfloor \frac{b_i-a_i -r_i+ 2\varepsilon_q}{2 \varepsilon_q} \right \rfloor \in \mathbb{Z}$. Then we can take 
\[
\beta_{i,j} = b_i - 2 \varepsilon_q (j-1)  \in B_i 
\]
for $j=1,2, \dots , c_i$.
Note that
$
\beta_{i,{j}}-\beta_{i,{j-1}} = 2\varepsilon_q
$ for $j=0,1,2, \dots , c_i$.
Thus the  half open intervals $ I_{\beta_{i,1}, \varepsilon_q},\dots , I_{\beta_{i,c_i}, \varepsilon_q} $ of length $2 \varepsilon_q$ are mutually disjoint.
Since $\beta_{i,j} \in B_i \subseteq A'$ we have
$\vert I_{\beta_{i,j}, \varepsilon_q} \cap  X_q \vert \ge \alpha$
for $j=1,2, \dots , c_i$ and 
\begin{displaymath}
\begin{split}
\vert B_i \cap X_q \vert 
&\ge
  \vert I_{\beta_{i,1}, \varepsilon_q} \cap  X_q \vert +  \dots+  \vert I_{\beta_{i,c_i}, \varepsilon_q} \cap  X_q \vert\\ 
& \ge \alpha c_i\\
&=	   \left \lfloor \frac{b_i-a_i -r_i + 2\varepsilon_q}{2 \varepsilon_q} \right \rfloor\alpha \\
& >  \left(\frac{b_i-a_i-r_i}{2 \varepsilon_q} \right) \alpha.
\end{split}
\end{displaymath}

\textbf{Case 1.} $m \in A$ and $M \in A$.\\
 For this case  $\mu_{\text{Bor}}(A')=\mu_{\text{Bor}}(A)+2\varepsilon_q$ and 
\begin{displaymath}
\begin{split}
	  n & \ge\vert A' \cap X_q \vert \\
&= \vert (B_1 \cup \dots \cup B_t) \cap X_q \vert\\
&=\vert B_1 \cap X_q \vert + \dots +\vert B_t \cap X_q \vert \\
&> \sum\limits_{i=1}^t \left(\frac{b_i-a_i-r_i}{2 \varepsilon_q} \right) \alpha \\
&=\frac{\alpha}{2\varepsilon_q} \mu_{\text{Bor}}(A') 
-\frac{r_1+\dots+r_t}{2\varepsilon_q}\alpha.
\end{split}
\end{displaymath}
Letting $r_1,\dots,r_t \rightarrow 0$, we have 
\[ n \ge \frac{\alpha}{2\varepsilon_q} \mu_{\text{Bor}}(A') \]
and 
\begin{displaymath}
\begin{split}
	  \mu_{\text{Bor}}(A)
&= \mu_{\text{Bor}}(A')-2\varepsilon_q\\
&\le \frac{2\varepsilon_q}{\alpha}n-2\varepsilon_q \\
&=\frac{l_q}{\alpha} \frac{n}{n-1}-2\varepsilon_q  \\
&=\frac{l_q}{\alpha} \left( \frac{n-\alpha}{n-1}  \right)\\
&<\frac{l_q}{\alpha} .
\end{split}
\end{displaymath}

\textbf{Case 2.}  ($m\in A$ and $M \notin A$) or ($m  \notin A$ and $M \in A$). \\
 For this case we have $\mu_{\text{Bor}}(A')=\mu_{\text{Bor}}(A)+\varepsilon_q$ and  $\vert A' \cap X_q \vert \le n-1$.
 Similarly, we have
\[
n-1 \ge \frac{\alpha}{2\varepsilon_q} \mu_{\text{Bor}}(A').
\]
Then 
\begin{displaymath}
\begin{split}
	  	  \mu_{\text{Bor}}(A)
&< \mu_{\text{Bor}}(A') \\
&\le \frac{2\varepsilon_q}{\alpha}(n-1)\\
&=\frac{l_q}{\alpha} .
\end{split}
\end{displaymath}

\textbf{Case 3.} $m  \notin A$ and $M \notin A$. \\
For this case, $\vert A' \cap X_q \vert \le n-2$ and we have
\[
n-2 \ge \frac{\alpha}{2\varepsilon_q} \mu_{\text{Bor}}(A').
\]
Then
\begin{displaymath}
\begin{split}
	  	  \mu_{\text{Bor}}(A)
&\le \mu_{\text{Bor}}(A') \\
&\le \frac{2\varepsilon_q}{\alpha}(n-2)\\
&< \frac{2\varepsilon_q}{\alpha}(n-1)\\
&=\frac{l_q}{\alpha} .
\end{split}
\end{displaymath}
\end{proof}

\begin{corollary}
The probability that we repeat \textbf{Stage 1} is less than $\frac{1}{\alpha}$.
\end{corollary}
\begin{proof}
Notice the following equivalence  
\[
\text{Repeat \textbf{Stage 1}}\;  \Leftrightarrow \;
\vert I_{(p,\varepsilon_q)} \cap X_q \vert \ge \alpha  \text{ in \textbf{Stage 2}}.
\]
By \textbf{Theorem \ref{thm1}}, we have
\begin{displaymath} 
\begin{split}
& \text{Prob}(\vert I_{(p,\varepsilon_q)} \cap X_q \vert \ge \alpha)\\ 
=&  \frac{\mu_{\text{Bor}}
(\{p \in [\min X_q ,\max X_q] : \vert I_{(p,\varepsilon_q)} \cap X_q \vert \ge \alpha \} 
}{\mu_{\text{Bor}}( [ \min X_q , \max X_q])}  \\
<& \frac{l_q}{\alpha}\frac{1}{l_q} \\
=& \frac{1}{\alpha}.
\end{split}
\end{displaymath}
\end{proof}

\begin{remark}\label{remark7}
\;
\begin{enumerate}[label=\roman*)]
\item The probability that Algorithm \ref{Density-Aware Partitioning_algorithm} terminates within $N-1$ iterations is {\it at least}
 $1 - \left( \frac{1}{\alpha} \right)^N$ and it is an increasing function with respect to $\alpha$ and $N$.
\item For $\alpha=2$, $\left( \frac{1}{2} \right)^6$ is an upper bound of the probability that we repeat \textbf{Stage 1} 6 times.
Therefore, $1-\left( \frac{1}{2} \right)^6=0.984375$ refers to a lower bound of the probability that the iteration  terminates within $5$ trials (Table \ref{table3}), and this probability further increases for $\alpha \geq 3$. Therefore, we need not concern ourselves with infinite loops.

\begin{table}[ht]
\centering
\begin{tabular}{|c|c|c|c|c|c|c|c|c|}
\hline
$N$& 1& 2& 3 &  4 & 5 & 6\\
\hline
$1 - \left( \frac{1}{2} \right)^N$& 0.50 & 0.75 & 0.88&  0.94& 0.97 & 0.98\\
\hline	
\end{tabular}
\caption{Lower bounds for the probability that the iteration  terminates within $N-1$ trials with $\alpha = 2$}\label{table3}

\end{table}
\end{enumerate}
\end{remark}

Example \ref{ex5} shows that the upper bound $\frac{l_q}{\alpha}$ in \textbf{Theorem \ref{thm1}} is \textit{sharp}. 
\begin{example}
Let $\alpha \ge2 $. Fix an integer $k$. For each $i=1,2,\dots, \alpha$, define
\[
L_i = \left  \{ (j,i) \mid j=1,2, \dots ,k  \right\}.
\]
Take  
\[ X= L_1 \cup \dots \cup L_\alpha  \cup \{ (0,0) , (\alpha k +1,0)\}\subseteq \mathbb{R}^2.\]
Then $X$ has $n=\alpha  k  +2 $ elements.
Suppose that the dimension cut is $q=1$. Note that 
\[
\varepsilon_1 = \frac{l_1}{2(n-1)}=\frac{\alpha k + 1}{2(\alpha k +2 -1)}=\frac{1}{2}.
\]
Then every point in 
$
X_1=\{0,1,2,\dots,k, \alpha k +1\}
$
except $0$ and  $\alpha k + 1$  has $\alpha$ duplicates.
Let 
 \[A=\{p \in [0 , \alpha k +1] : \vert I_{(p,\varepsilon_1)} \cap X_1 \vert \ge \alpha  \}.\]
Then we have 
$
	\mu_{\text{Bor}}(A)=k \times 2\varepsilon_1 =k.
$
 On the other hand, by  \textbf{Theorem \ref{thm1}}, we have an upper bound of $\mu_{\text{Bor}}(A)$:
\[
\frac{l_1}{\alpha}=\frac{\alpha k + 1}{\alpha} =k + \frac{1}{\alpha}.\]
Note that 
\[ k + \frac{1}{\alpha} \rightarrow \mu_{\text{Bor}}(A) \; \text{ as } \; \alpha \rightarrow \infty.\]
This shows  that the upper bound %of $\mu_{\text{Bor}}(A)$ 
converges to $\mu_{\text{Bor}}(A)$.
\label{ex5}\end{example}

\subsection{Convergence and Complexity}
Both the IF and RRCF algorithms have similar time complexities  $O(mn\log n)$ where $n$ is the data size and $m$ is the number of trees \cite{4781136,guha2016robust}. The density-aware partitioning affects only the selection of split values, which are repeated. The number of repetitions, denoted as $N$, does not depend on the data size. In fact, $N$ is bounded above by $M$ where $M$ follows a geometric distribution with parameter $\frac{1}{\alpha}$ (Remark \ref{remark7}). Therefore, the expected number of repetitions is $\alpha$, which is independent of the data size.

Now, we will observe that the anomaly scores in the WIF and WRCF do not converge to the same limit as in IF and RRCF, respectively, as demonstrated in Example \ref{ex4}.
\begin{example}\label{ex4}
Let $X = \{a, b, c, d\} \subseteq \mathbb{R}^2$ be a set of four points defined as follows:
\[ a=(0,0), \; b=(1,0), \; c=(6,0), \; d=(7,0).\]

Since the $y$-coordinates of all points in $X$ are zero, we will only consider $P|_{Q=1}$. Table \ref{table4} shows the anomaly score of each algorithm. Note that the anomaly scores of all points in $X$ with the WIF and WRCF are equal, as Algorithm \ref{Density-Aware Partitioning_algorithm} allows building only tree $T_1$ in Fig. \ref{fig12}.

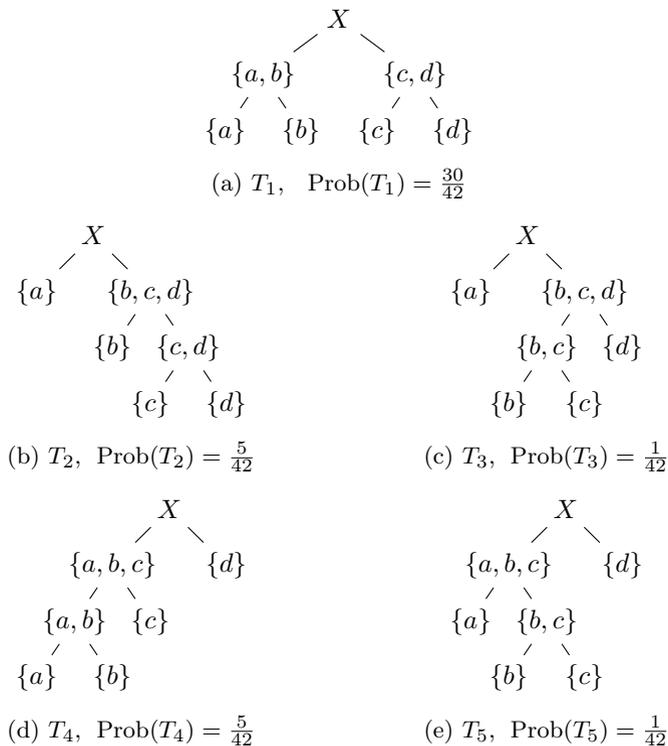
\begin{figure}[h]
\centering
\begin{subfigure}{0.45\textwidth}
\centering
\begin{tikzpicture}[scale=0.5] % Adjust the scale as needed
\node  {$X$}
[
level 1/.style = {sibling distance = 4cm},
level 2/.style = {sibling distance = 2cm}
]
    child {node {$\{a,b\}$}
        child{ node {$\{a\}$}}
        child{ node {$\{b\}$}}}
    child {node {$\{ c,d\}$}
        child{ node {$\{c\}$}}
        child{ node {$\{d\}$}}}
;
\end{tikzpicture}
\caption{$T_1$, \; $\text{Prob}(T_1) = \frac{30}{42}$}
\end{subfigure}\\
\begin{subfigure}{0.45\textwidth}
\centering
\begin{tikzpicture}[scale=0.5]
\node  {$X$}
[
level 1/.style = {sibling distance = 3cm},
level 2/.style = {sibling distance = 2cm}
]
    child {node {$\{a\}$}}
    child {node {$\{ b, c,d\}$}
        child{ node {$\{b\}$}}
        child{ node {$\{c,d\}$}
            child{ node {$\{c\}$}}
            child{ node {$\{d\}$}}}}
   ;
\end{tikzpicture}
\caption{$T_2$,\; $\text{Prob}(T_2) = \frac{5}{42}$}
\end{subfigure}
\begin{subfigure}{0.45\textwidth}
\centering
\begin{tikzpicture}[scale=0.5]
\node  {$X$}
[
level 1/.style = {sibling distance = 3cm},
level 2/.style = {sibling distance = 2cm}
]
    child {node {$\{a\}$}}
    child {node {$\{ b, c,d\}$}
        child{ node {$\{b,c\}$}
            child{ node {$\{b\}$}}
            child{ node {$\{c\}$}}}
        child{ node {$\{d\}$}}}
   ;
\end{tikzpicture}
\caption{$T_3$,\; $\text{Prob}(T_3) = \frac{1}{42}$}
\end{subfigure}
\begin{subfigure}{0.45\textwidth}
\centering
\begin{tikzpicture}[scale=0.5]
\node  {$X$}
[
level 1/.style = {sibling distance = 3cm},
level 2/.style = {sibling distance = 2cm}
]
    child {node {$\{a,b,c\}$}
        child{node{ $ \{a,b \}$}
                            child{node{ $ \{a \}$}}
                            child{node{ $ \{b \}$}}
                }
        child{node{ $ \{c\}$}}}
    child {node {$\{ d\}$}}
   ;
\end{tikzpicture}
\caption{$T_4$,\; $\text{Prob}(T_4) = \frac{5}{42}$}
\end{subfigure}
\begin{subfigure}{0.45\textwidth}
\centering
\begin{tikzpicture}[scale=0.5]
\node  {$X$}
[
level 1/.style = {sibling distance = 3cm},
level 2/.style = {sibling distance = 2cm}
]
    child {node {$\{a,b,c\}$}
        child{node{ $ \{a \}$}}
        child{node{ $ \{b,c\}$}
                        child{node{ $ \{b\}$}}
                    child{node{ $ \{c \}$}} } }
    child {node {$\{ d\}$}}
   ;
\end{tikzpicture}
\caption{$T_5$,\; $\text{Prob}(T_5) = \frac{1}{42}$}
\end{subfigure}
\caption{The graphs of $T_i$ with Prob($T_i$) (Example \ref{ex4})}\label{fig12}
\end{figure}

\noindent \textbf{Ordinary Partitioning:}
In both the IF and RRCF algorithms, there are five possible trees denoted as $T_i$. The $\text{Prob}(T_i)$ represents the probability of obtaining $T_i$ during the tree-building process based on Algorithm \ref{Ordinary Partitioning_algorithm}, as illustrated in Fig. \ref{fig12}.

\begin{table}
    \centering
    
    \begin{tabular}{|c|c|c|c|c|c|}
        \hline
        $x\in X$ & $a$ & $b$ & $c$ & $d$ \\
        \hline
        \textbf{RRCF} & $1.31$ & 1.12 & 1.12 & 1.31 \\
        \hline
        \textbf{WRCF} & $1$ & $1$ & $1$ & $1$ \\
        \hline
        \textbf{IF} & $0.48$ & $0.45$ & $0.45$ & $0.48$\\
        \hline
        \textbf{WIF} & $0.5$ & $0.5$ & $0.5$ & $0.5$\\
        \hline
    \end{tabular}	
    \caption{Anomaly scores for the RRCF, WRCF, IF, and WIF}
    \label{table4}
\end{table}

\noindent \textbf{Density-Aware Partitioning:} Choose $\alpha = 2$ in Algorithm \ref{Density-Aware Partitioning_algorithm}. Note that the radius $\varepsilon_{1}$ of $X_1$  is $\frac{7}{6}$ and so
\[
\vert I_{p,\varepsilon_{1}} \cap X_1 \vert < 2 
\;\Leftrightarrow \;  p \in \left(\frac{7}{6}, \frac{35}{6}  \right]
\]
but $\left(\frac{7}{6}, \frac{35}{6}  \right] \subseteq (1,6)$. So, we have the tree $T_1$ (Fig. \ref{fig12}) with the probability $1 - \left(\frac{1}{3} \right)^{N+1}$, where $N$ is  the number of trials we repeat \text{Stage 1} in Algorithm \ref{Density-Aware Partitioning_algorithm}. In other words, with the WIF and WRCF algorithms, we always obtain the tree $T_1$.
\end{example}

\section{Examples}\label{Examples}
In this section we conduct a comparative analysis of the RRCF and WRCF algorithms across two distinct domains: time series data (\ref{time series data}) and Euclidean data (\ref{euclidean data}). Additionally, we evaluate the IF, WIF, RRCF, and WRCF algorithms using benchmark data sets (\ref{benchmark data sets}). 
\subsection{Time series Data}\label{time series data}

\subsubsection{Synthetic data}
\begin{figure}[h]
    \centering
    \includegraphics[scale=0.4]{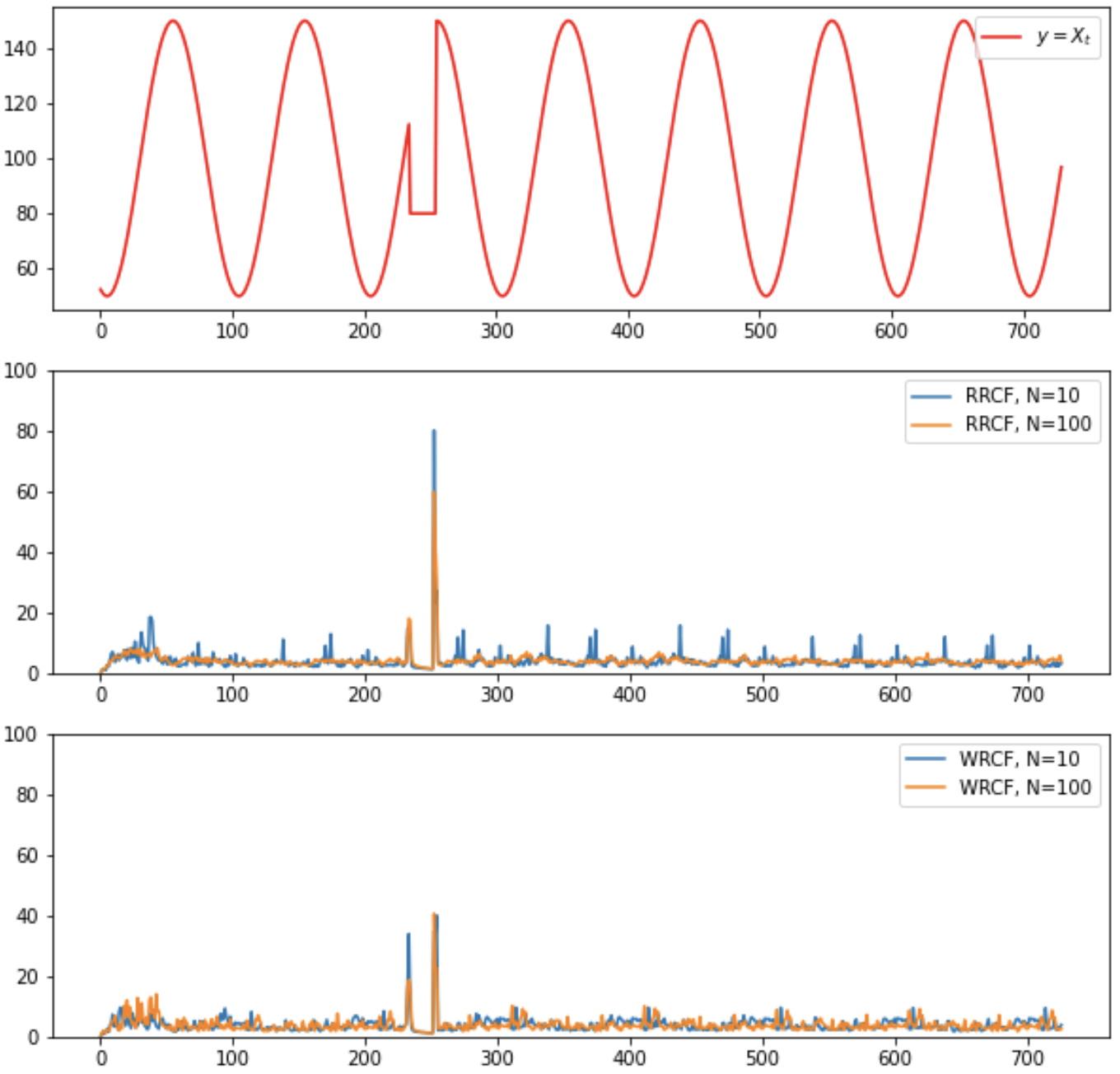}
    \caption{$X_t$ and $\textbf{Avg-CODISP}$ of RRCF and WRCF}
    \label{fig13}
\end{figure}

%Suppose we have the following 
Consider the following time series data $ \{X_t\}$ 
with $t=1,2,\dots, 730$,
\[
X_t =
\begin{cases}
80 & \text{if $t=235, 234, \dots, 254$}\\
	50 \sin (\frac{2 \pi}{50} (t-30)) & \text{elsewhere}
\end{cases}.
\]
With the shingle, window, and forest sizes of $4,256$, and $40$, respectively,
the RRCF and  WRCF algorithms yield Fig. \ref{fig13}.
It shows that both the RRCF and WRCF algorithms detect the desired anomalies at $t = 235$ and $254$ successfully. The figure also shows that the WRCF algorithm converges faster than the RRCF (shown in blue ($N = 10$) and orange ($N = 100$) lines). 
\FloatBarrier

\subsubsection{GDP growth rate data}
\begin{figure}[h]
\centering
			       \includegraphics[scale=0.4]{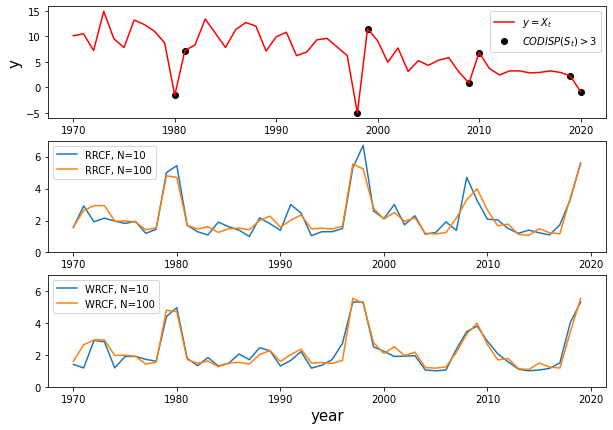}
\caption{Korea GDP versus year (top)  and $\textbf{Avg-CODISP}$s of RRCF (middle)  and WRCF (bottom)}
\label{fig14}
\end{figure}

Gross Domestic Product (GDP) is a monetary measure of the market value of all final goods and services produced in a specific period. The GDP growth rate for each period is defined as the ratio of the difference between GDP values from the current period to the next period to the GDP value from the earlier period, multiplied by 100.
\[
\text{GDP Growth Rate} = \left( \frac{\text{GDP}_{\text{current period}} - \text{GDP}_{\text{earlier period}}}{\text{GDP}_{\text{earlier period}}} \right) \times 100
\]
When economic shocks, such as the global oil shock (1980) and the global financial crisis (2009), hit the world, the GDP of each country plunged. We apply the RRCF and WRCF algorithms to the GDP growth rate data of Korea \cite{ecos} with the following parameters: shingle size = 2, window size = 8, and the number of iterations = $N$. The results are shown in Fig. \ref{fig14}. The top figure displays the anomalies detected where $\textbf{CODISP} > 3$ with both the RRCF and WRCF algorithms. The middle and bottom figures show the \textbf{Avg-CODISP}s of the RRCF (middle) and WRCF (bottom) for $N = 10$ (blue) and $N = 100$ (orange). As shown in those figures, the WRCF algorithm yields faster convergence than the RRCF.

Table \ref{table5}  shows the CODISPs of the RRCF and WRCF with $N = 100$. As seen in the table, both algorithms yield similar CODIPS values and capture all the desired economic shocks in the data. 

\begin{table}
\centering
\begin{tabular}{ |c|c|c|c| c| c| } 
\hline
Year & Event   & RRCF & WRCF \\
\hline
1980& Oil Shock & 4.4 &4.4 \\ 
\hline
1981& Oil Shock Recovery& 4.5 &5 \\ 
\hline
1998& IMF Credit Crisis & 5.4  & 5.3\\
\hline
1999& IMF Credit Crisis Recovery  & 5.3  &5.3\\
\hline
2009& Global Finance Crisis   & 3.4  & 3.5\\
\hline
2010& Global Finance Crisis Recovery   & 3.9  & 3.8\\
\hline
2019& COVID-19  &  3.7 & 4  \\
\hline
2020& COVID-19  & 5.6  &5.3 \\
\hline
 \end{tabular}
\caption{Economic shocks in Korea and \textbf{CODISP}s of RRCF and WRCF}\label{table5}	
\end{table}

\FloatBarrier

\subsubsection{NewYork Taxi data}
We also use the  taxi ridership data from the NYC Taxi Commission \cite{NYtaxi}, where the total number of passengers is aggregated over a 30-minute time window \cite{guha2016robust}. We employ  the following parameters: shingle size = $48$ (a day), sample size = $1000$, and total iteration number = $N$. The deleting and inserting method (Section \ref{RRCFtime}) is used.

\begin{figure}[h]
\centering
\includegraphics[scale= 0.213]{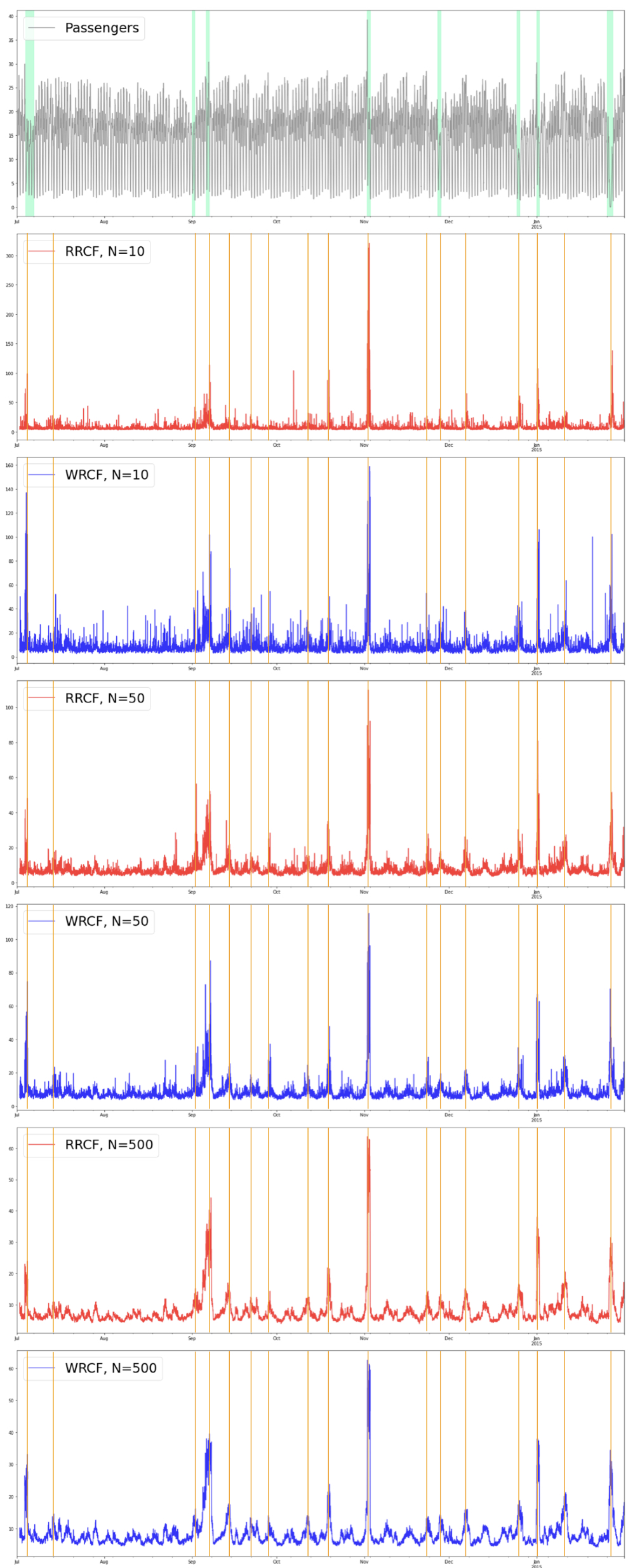}	
\caption{\textbf{Avg-CODISP}s of RRCF (red) and WRCF (blue)}
\label{fig15}
\end{figure}

The top figure of Fig. \ref{fig15} displays the given taxi ridership data, and the subsequent figures depict the graphs of \textbf{Avg-CODISP}s with the RRCF (red) and WRCF (blue) algorithms for $N=10, 50, 500$. The top figure shows the taxi ridership data, with special days (such as holidays) highlighted in green, during which the anomalous behaviors of taxi ridership occurred. We consider \textbf{Avg-CODISP}s with $N=500$ as the limit values for both algorithms, regarding days with $\textbf{Avg-CODISP}>14$ as anomalies for both algorithms. The bottom two figures show a total of $17$ anomalies identified at the limit ($N = 500$), highlighted with orange lines in each \textbf{Avg-CODISP} figure as a reference. We consider these 17 limit anomalies as the {\it desired} anomalies that we are looking for. Any other anomalies detected are considered {\it false} anomalies. The actual meaning of those anomalies in the physical sense is not of interest; we are mainly concerned with the convergent behaviors of each method.

To assess how each method detects the desired anomalies as $N$ changes, we computed \textbf{Avg-CODISP} with $N = 10$ and $N = 50$. As shown in the second figure, the RRCF algorithm identified 8 anomalies out of the 17 desired anomalies. However, the third figure shows that the WRCF found all 17 anomalies. Thus, the WRCF algorithm seems to identify all the desired anomalies faster than the RRCF algorithm. Figures with $N = 10$ also reveal that the WRCF found more false anomalies than the RRCF. Figures with $N = 50$ show that the RRCF algorithm found more desired anomalies than with $N = 10$, but not all 17, while the WRCF algorithm still identified all 17 desired anomalies and found fewer false anomalies than with $N = 10$. This experiment also implies that the WRCF algorithm provides faster convergent results than the RRCF algorithm.

\FloatBarrier

\subsection{Euclidean Data}\label{euclidean data}
\subsubsection{10 points with one cluster}
\begin{figure}[h]
\centering
\includegraphics[scale=0.4]{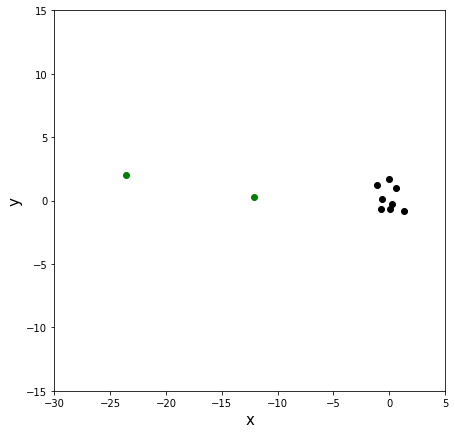}
\caption{The graph of $X$, $\mu(X) = 0.55$}\label{fig16}
\end{figure}

Suppose we have the following data set $X = \{a, b, c, d, e, f, g, h, i, j\} \subseteq \mathbb{R}^2$, depicted in Fig. \ref{fig16}, where
\begin{displaymath}
\begin{split}
a & = (-23.6, -2), \quad b=(-12.1, 0.3), \quad c=(0, 1.7), \\
d & = (-1.1, 1.2), \quad e = (0.6, 1), \quad f = (0.3, -0.3), \\
g & = (0.1, -0.7), \quad h = (1.3, -0.8), \quad i = (-0.7, -0.7), \quad j = (-0.6, 0.1).
\end{split}
\end{displaymath}

Tables \ref{table6} and \ref{table7} show   \textbf{Avg-CODISP} of each point $x\in X$ where $N$ is the number of iterations.
\begin{table}
\begin{center}
\begin{tabular}{ |c|c|c|c|c|c|c|c|c|c|c|c|} 
\hline
        $N$           &  a & b & c &d & e& f&g&h&i&j\\
\hline
$1$ & 4 & 4 &2&   1.6 & 1&
		 2& 1& 4& 3 & 1\\
\hline
$2$ &  6.5 & 6 & 1.8  & 1.8    & 1.3& 
				1.5& 1& 3& 2.3& 1.6\\
\hline
$3$ &  7.3 & 6.6 & 1.8  & 2.2 & 1.2 & 
				1.3 & 1& 3& 1.9& 1.7\\
\hline
$4$ &  7.8 & 2 & 3.4  & 1.2 & 1.3 & 
				1.3 & 1& 3.3& 1.7& 1.9\\
\hline
$5$ &  8 & 7.2 & 2.8  & 3.1 & 1.1 & 
				1.2 & 1& 4& 1.5& 1.9\\
\hline
$10$ &  7.5 & 6.5 & 2.1  & 3.1 & 1.1 & 
				1.2 & 1& 4& 1.5& 1.9\\
\hline
$10000$ &  6.1 & 5.2 & 2.5  & 2.3 & 1.6 & 
				1.4 & 1.2& 3.3& 1.7& 2.0\\
\hline
\end{tabular}
\end{center}
\caption{RRCF: \textbf{Avg-CODISP}s for each point}\label{table6}
\end{table}

\begin{table}
\begin{center}
\begin{tabular}{ |c|c|c|c|c|c|c|c|c|c|c|c|} 
\hline
        $ N$          &  a & b & c &d & e& f&g&h&i&j\\
\hline
$1$ & 9 & 8 & 7&   6 & 2&
		 1& 1& 5& 2 & 4\\
\hline
$2$ &  9 & 8 & 4.2  & 4    & 3& 
				1& 1& 6& 2& 2.7\\
\hline
$3$ &  7.3 & 6.7 & 3.8  & 3 & 2.3 & 
				1 & 1& 5& 2& 2\\
\hline
$4$ &  7.8 & 7 & 4.6  & 2.8 &2 & 
				1.2 & 1.2& 4.4& 1.8& 2.5\\
\hline
$5$ &  7 & 6.4 & 4.1  & 2.6 & 1.8 & 
				1.1 & 1.1& 4.9& 1.6& 3.2\\
\hline
$10$ &  6 & 5.7 & 3.2  & 2.1 & 1.8 & 
				1.3 & 1.1& 3.8& 1.8& 2.4\\
\hline
$10000$ &  6.4 & 5.8 & 2.7  & 2.4 & 1.7 & 
				1.4 & 1.2& 3.4& 1.6& 2.0\\
\hline
\end{tabular}
\end{center}
\caption{WRCF: \textbf{Avg-CODISP}s for each point} \label{table7} 
\end{table}

Note that points ${a},  {b}$ have
higher CODISP values than the others in both algorithms.
The values
\begin{displaymath}
\begin{split}
	  \sum\limits_{x \in X} \left( \left\vert \textbf{Avg-CODISP}(x,10000) -\textbf{Avg-CODISP}(x,10) \right\vert  \right)
\end{split}
\end{displaymath}
of Tables  \ref{table6} and \ref{table7} are $5.9$ and $2.6$, respectively.
This shows that the WRCF algorithm converges to $\textbf{Avg-CODISP}(x,10000)$ faster than the RRCF.
\FloatBarrier

\subsubsection{Normally distributed points}
Let $\mathbb{X}$ be a set of 100 points in $\mathbb{R}^2$ where
each $(x,y) \in \mathbb{X}$ is obtained from the random variable $(X,Y)$ such that
both $X$ and $Y$  are normally distributed. 
The density measure $\mu(X)$ is 0.06. 
One could assume that the likelihood of a point being an anomaly increases as its distance from the origin (0,0) grows. We run the RRCF and WRCF algorithms with the $\text{sample size}=10$ and $ \text{the number of iterations }=N$.

\begin{figure}[h]
    \centering
    \begin{subfigure}[b]{0.45\textwidth}
        \includegraphics[scale=0.35]{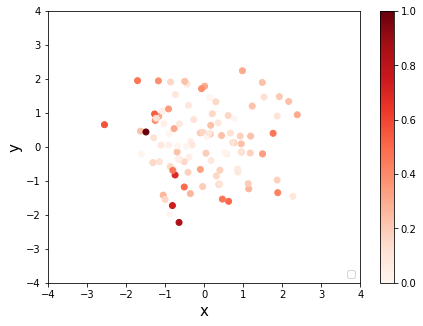}
        \caption{$N = 20$}
    \end{subfigure}
    \begin{subfigure}[b]{0.45\textwidth}
        \includegraphics[scale=0.35]{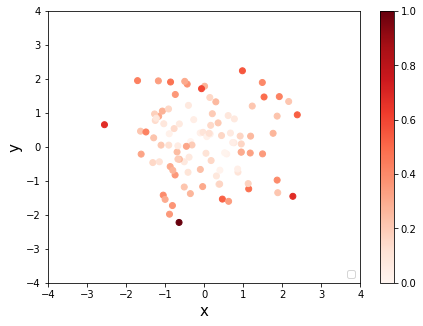}
        \caption{$N = 100$}
    \end{subfigure}
    \caption{\textbf{Avg-CODISP} of RRCF}
    \label{fig17}
\end{figure}

\begin{figure}[h]
    \centering
    \begin{subfigure}[b]{0.45\textwidth}
        \includegraphics[scale=0.35]{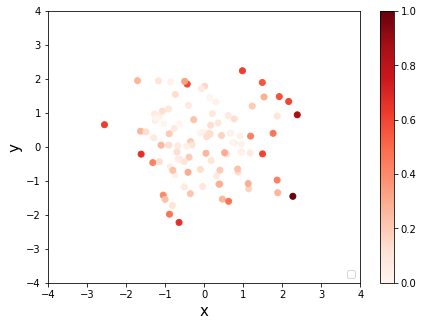}
        \caption{$N = 20$}
    \end{subfigure}
    \begin{subfigure}[b]{0.45\textwidth}
        \includegraphics[scale=0.35]{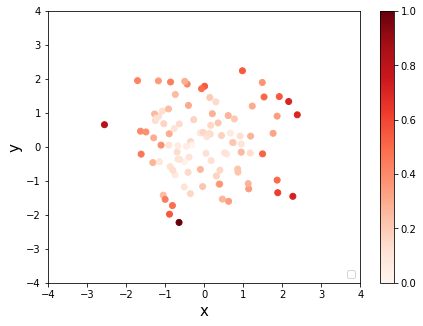}
        \caption{$N = 100$}
    \end{subfigure}
    \caption{\textbf{Avg-CODISP} of WRCF}
    \label{fig18}
\end{figure}

\noindent 
The color bars in Figs.  \ref{fig17} and \ref{fig18} depict the
 (normalized) \textbf{Avg-CODISP} values for the points in $X$.
\textcolor{black}
{When $N=100$}, both the  RRCF and  WRCF algorithms capture those points far from the origin as anomalies.
When \textcolor{black}
{$N=20$}, the RRCF algorithm does not capture anomalies well compared to the WRCF algorithm
since many points far from the origin have lower \textbf{CODISP} values with the RRCF.
To capture anomalies well enough, the  RRCF needs a larger value of $N$ than the  WRCF. 
\FloatBarrier

\subsubsection{Normally distributed points with anomaly clusters}
Let $X_1,X_2'$ and $X_3'$ be sets of $100,5$ and $5$ points in $\mathbb{R}^2$, respectively, where
each point in the sets is obtained from the random variable $(X, Y)$ such that
both $X$ and $Y$  are normally distributed. 
Denote
\begin{displaymath}
\begin{split}
	  X_2 =& X_2' -(5,5) = \{(x,y)-(5,5) \mid (x,y) \in X_2' \},\\
	  X_3 =&  X_3' +(5,5)= \{(x,y)+(5,5) \mid (x,y) \in X_3' \}
\end{split}
\end{displaymath}
and set $X=X_1 \cup X_2 \cup X_3$. The density measure $\mu(X)$ is 0.07. 
It may be presumed that $X_2$ and $X_3$ are clusters of anomalies.
\begin{figure}[h]
    \centering
    \begin{subfigure}[b]{0.45\textwidth}
        \includegraphics[scale=0.4]{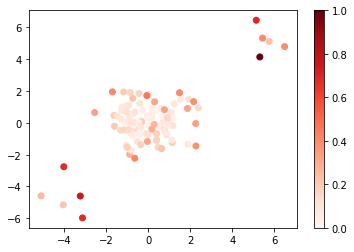}
        \caption{RRCF}
    \end{subfigure}
    \begin{subfigure}[b]{0.45\textwidth}
        \includegraphics[scale=0.4]{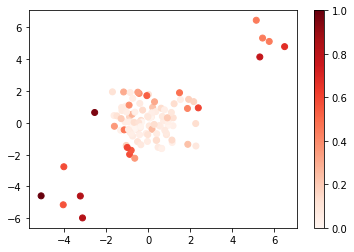}
        \caption{WRCF}
    \end{subfigure}
    \caption{\textbf{Avg-CODISP} of RRCF and WRCF}
    \label{fig19}
\end{figure}

\noindent The color bars in Fig. \ref{fig19} refer to
the normalized \textbf{Avg-CODISP}s for the points in $X$.
If we consider points with (normalized) $\textbf{Avg-CODISP}>0.5$ as anomalies,
the   RRCF captures 3 points in $X_2$ and 4 points in $X_3$ as anomalies 
but the  WRCF captures all the points in $X_2$ and $X_3$ as anomalies.
Moreover,  the  WRCF captures points that are far from the origin in $X_1$ as anomalies
whereas the   RRCF treats the whole $X_1$ as normal data.

\FloatBarrier

\subsubsection{Normally distributed clusters with noises}
Let $X_1$ be a set of two clusters where each cluster is obtained by multiplying $0.1$ by normally distributed 100 points with translation. Let $X_2$ be a set of normally distributed  40 points. Let $X=X_1 \cup X_2$.
\begin{figure}[h]
\centering
\includegraphics[scale=0.6]{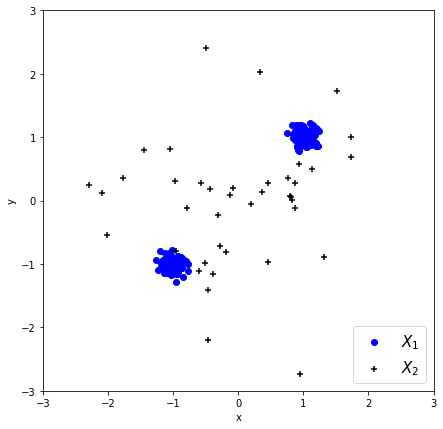}	
\caption{The graph of $X$, $\mu(X)=0.09$}
\label{fig20}
\end{figure}
In Fig. \ref{fig20}, the blue dots denote the data set of $X_1$ and the black crosses of $X_2$. As shown in the figure, $X_1$ is composed of two clusters while the elements of $X_2$ are not clustered but scattered. Thus apparently $X_2$ seems more like noise or anomalies compared to the clustered $X_1$. 
We apply the RRCF and WRCF algorithms with the  $\text{sample size}=20$ and $ \text{the number of iterations}=N$.
In Figs. \ref{fig21} and \ref{fig22}, the more red a point is, the higher \textbf{Avg-CODISP} it has.
Note that the RRCF algorithm cannot capture 3 points near $(-2,0)$ as anomalies with $N=10$ and they are regarded as anomalies with $N=100$. However, the  WRCF captures the 3 points as anomalies even with $N=10$.

\begin{figure}[h]
    \centering
    \begin{subfigure}[b]{0.45\textwidth}
        \includegraphics[scale=0.35]{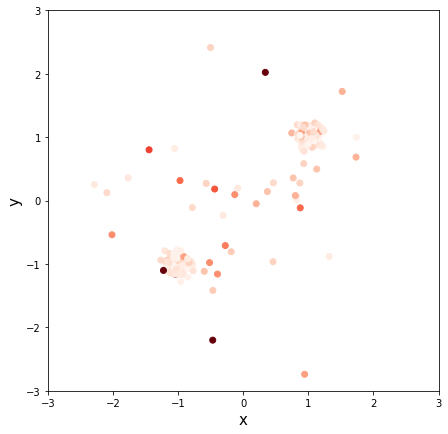}
        \caption{$N=10$}
    \end{subfigure}
    \begin{subfigure}[b]{0.45\textwidth}
        \includegraphics[scale=0.35]{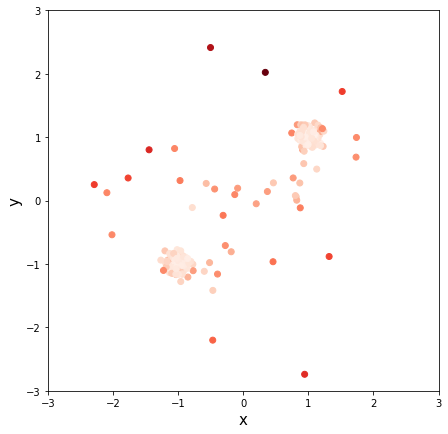}
        \caption{$N=100$}
    \end{subfigure}
    \caption{\textbf{Avg-CODISP} of RRCF}
    \label{fig21}
\end{figure}

\begin{figure}[h]
    \centering
    \begin{subfigure}[b]{0.45\textwidth}
        \includegraphics[scale=0.35]{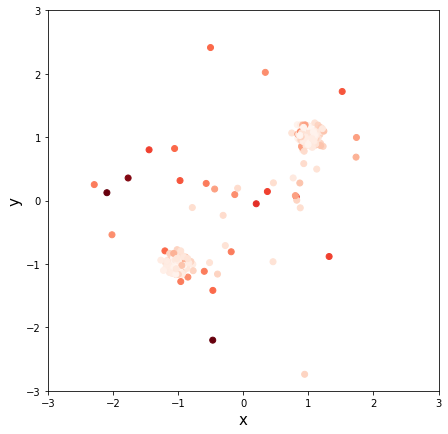}
        \caption{$N=10$}
    \end{subfigure}
    \begin{subfigure}[b]{0.45\textwidth}
        \includegraphics[scale=0.35]{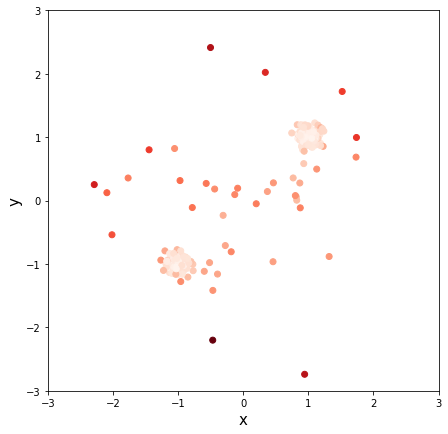}
        \caption{$N=100$}
    \end{subfigure}
    \caption{\textbf{Avg-CODISP} of WRCF}
    \label{fig22}
\end{figure}

\begin{figure}[h]
    \centering	 
    \begin{subfigure}{0.45\textwidth}
        \includegraphics[scale = 0.33]{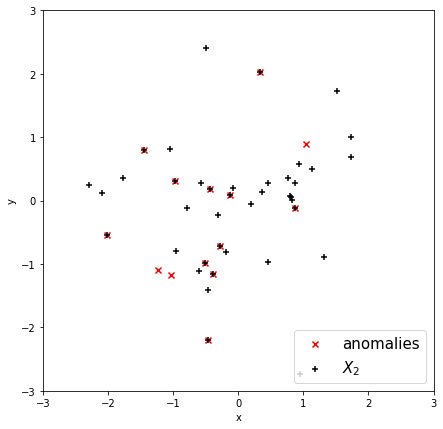}
        \caption{$N = 10$, $\vert X_2 \cap A \vert = 11$}
    \end{subfigure}
    \begin{subfigure}{0.45\textwidth}
        \includegraphics[scale = 0.33]{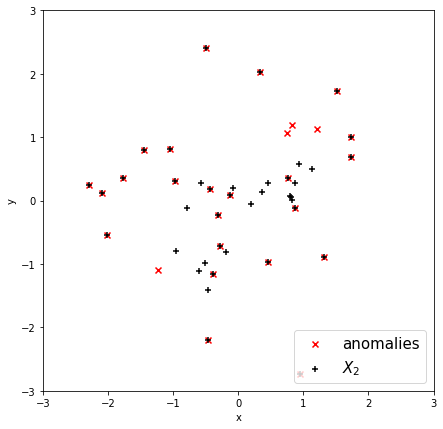}
        \caption{$N = 100$, $\vert X_2 \cap A \vert = 23$}
    \end{subfigure}
    \begin{subfigure}{0.45\textwidth}
        \includegraphics[scale = 0.33]{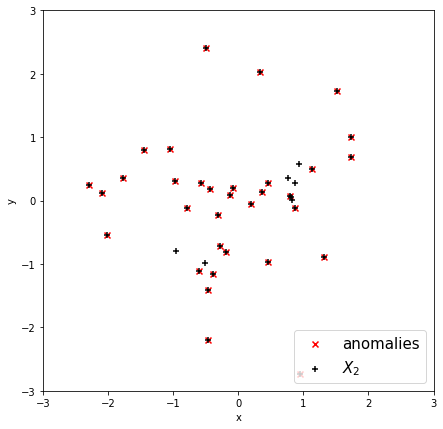}
        \caption{$N = 1000$, $\vert X_2 \cap A \vert = 33$}
    \end{subfigure}
    \begin{subfigure}{0.45\textwidth}
        \includegraphics[scale = 0.33]{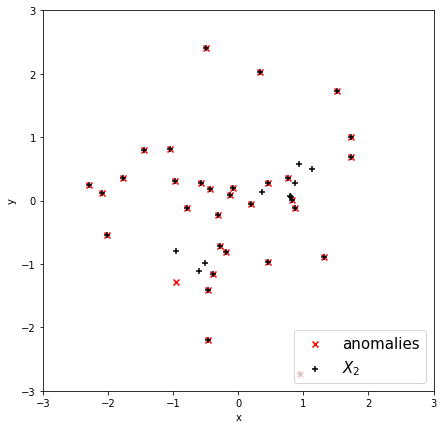}
        \caption{$N = 10000$, $\vert X_2 \cap A \vert = 31$}
    \end{subfigure}
    \caption{Detected anomalies of RRCF}
    \label{fig23}
\end{figure}

\begin{figure}[h]
    \centering	 
    \begin{subfigure}{0.45\textwidth}
        \includegraphics[scale = 0.33]{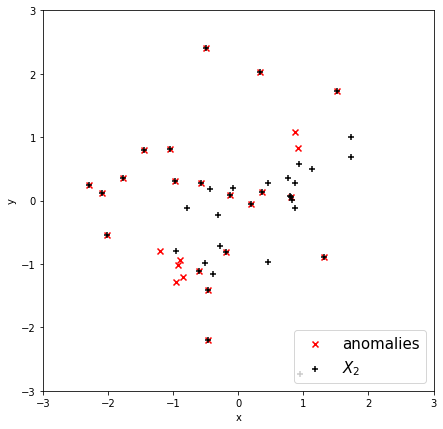}
        \caption{$N = 10$, $\vert X_2 \cap A \vert = 20$}
    \end{subfigure}
    \begin{subfigure}{0.45\textwidth}
        \includegraphics[scale = 0.33]{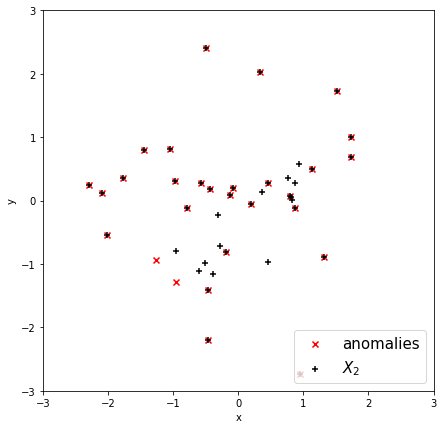}
        \caption{$N = 100$, $\vert X_2 \cap A \vert = 27$}
    \end{subfigure}
    \begin{subfigure}{0.45\textwidth}
        \includegraphics[scale = 0.33]{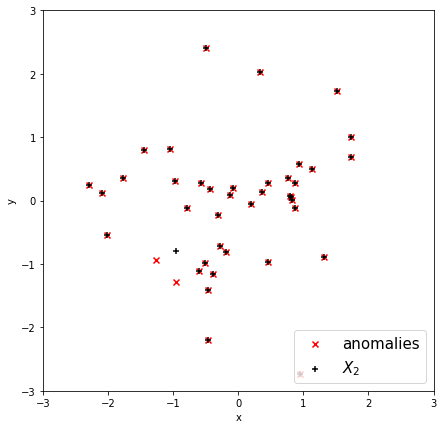}
        \caption{$N = 1000$, $\vert X_2 \cap A \vert = 39$}
    \end{subfigure}
    \begin{subfigure}{0.45\textwidth}
        \includegraphics[scale = 0.33]{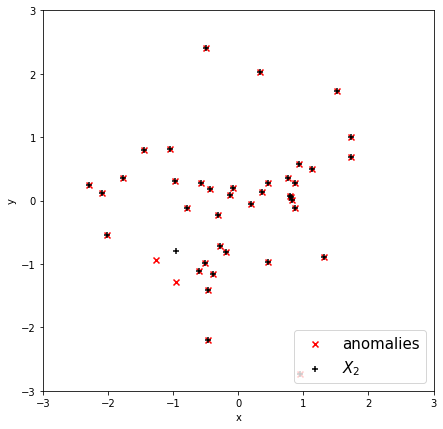}
        \caption{$N = 10000$, $\vert X_2 \cap A \vert = 39$}
    \end{subfigure}
    \caption{Detected anomalies of WRCF}
    \label{fig24}
\end{figure}

Let $A$ be the subset of $X$ whose points have the normalized $\text{Avg-CODISP}>0.35$.
One may think that $A$ is the set of detected anomalies whereas $X_2$ is the set of anomalies.
One may think that $\vert X_2 \cap A \vert$, which is the number of detected anomalies in $X_2$, measures how well the algorithm detects anomalies. In Figs. \ref{fig23} and \ref{fig24}, we observe that, for every value of $N$ used, the WRCF consistently exhibits a larger $\vert X_2 \cap A \vert$ compared to the RRCF.

\begin{figure}[h]
    \centering	 
    \begin{subfigure}{0.45\textwidth}
        \includegraphics[scale = 0.32]{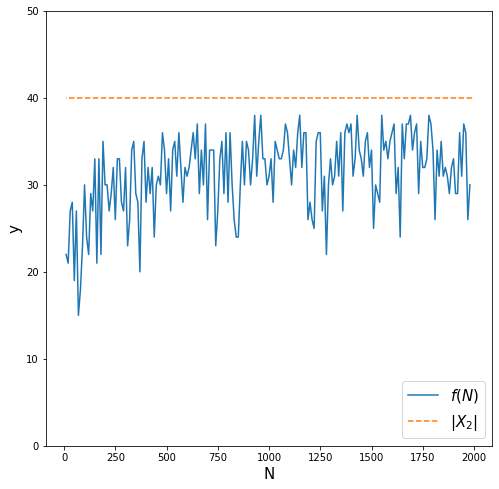}
        \caption{RRCF}
    \end{subfigure}
    \begin{subfigure}{0.45\textwidth}
        \includegraphics[scale = 0.32]{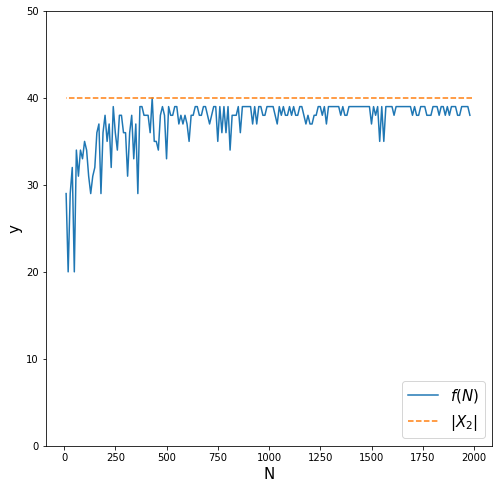}
        \caption{WRCF}
    \end{subfigure}
    \caption{$y=f(N)$}
    \label{fig25}
\end{figure}

\begin{figure}[h]
    \centering	 
    \begin{subfigure}{0.45\textwidth}
        \includegraphics[scale = 0.32]{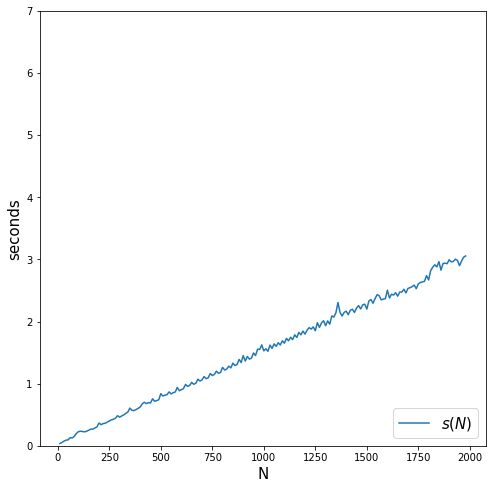}
        \caption{RRCF}
    \end{subfigure}
    \begin{subfigure}{0.45\textwidth}
        \includegraphics[scale = 0.32]{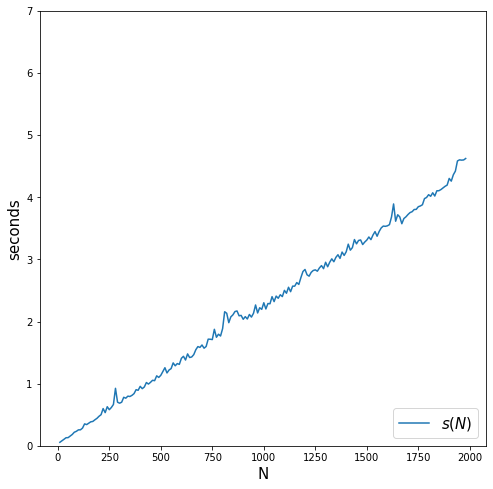}
        \caption{WRCF}
    \end{subfigure}
    \caption{$y=s(N)$}
    \label{fig26}
\end{figure}

In order to understand qualitatively how the number of the detected anomalies behaves with $N$, 
we let $f(N)=\vert X_2 \cap A \vert$, the number of the detected anomalies. Note that the desired number of anomalies in $X$ is $\vert X_2 \vert =40$.  Figure  \ref{fig25} shows  the graph of $f(N)$ 
versus $N$  with the RRCF (left) and WRCF (right) algorithms where $N=10,20,30, \dots, 2000$.  In the figure, the orange dotted line indicates the desired number of anomalies, i.e., $\vert X_2 \vert =40$.  The number of the detected anomalies, $f(N)$,  is depicted in blue. As shown in Fig. \ref{fig25}, $f(N)$ of the RRCF fluctuates highly more than the WRCF. For the WRCF, the magnitude of the fluctuation decreases with $N$ while the magnitude of the fluctuation with the RRCF does not change even though $N$ increases. According to the pattern of $f(N)$ of the RRCF (left), it seems that  $f(N)$ does not necessarily reach the desired number of anomalies even though $N$ is highly large.  We clearly see that the WRCF is more effective and stable than the RRCF as $N$ increases. $f(N)$ with the WRCF reaches the desired number of $\vert X_2 \vert =40$ much faster than the RRCF. 

To check the computational complexity of both algorithms for this problem, we let $s(N)$ be the running time of each algorithm in seconds. Figure   \ref{fig26} shows the running time $s(N)$ versus $N$ of the RRCF (left) and WRCF (right) algorithms. First of all, as expected, we observe that $s(N)$ is linear with respect to $N$ for both algorithms. The WRCF has a slightly steeper slope of $s(N)$ than the RRCF -- the slope of $s(N)$ of the WRCF is about $1.5$ times larger than that of the  RRCF. When $N$ is small, the difference is not significant.

\FloatBarrier

\subsection{Benchmark data sets}\label{benchmark data sets}

The performance of the IF, WIF, RRCF, and WRCF algorithms is   evaluated across eight benchmark datasets. 
The details of every dataset used in the experiments are outlined in Table \ref{table8}. The AUC score, commonly  used to evaluate the performance of a binary classification model, is measured on the $y$-axis, while the $x$-axis represents the tree size for each dataset in Table \ref{table8}.

\begin{figure}[h] 
    \centering
    \includegraphics[scale=0.2]{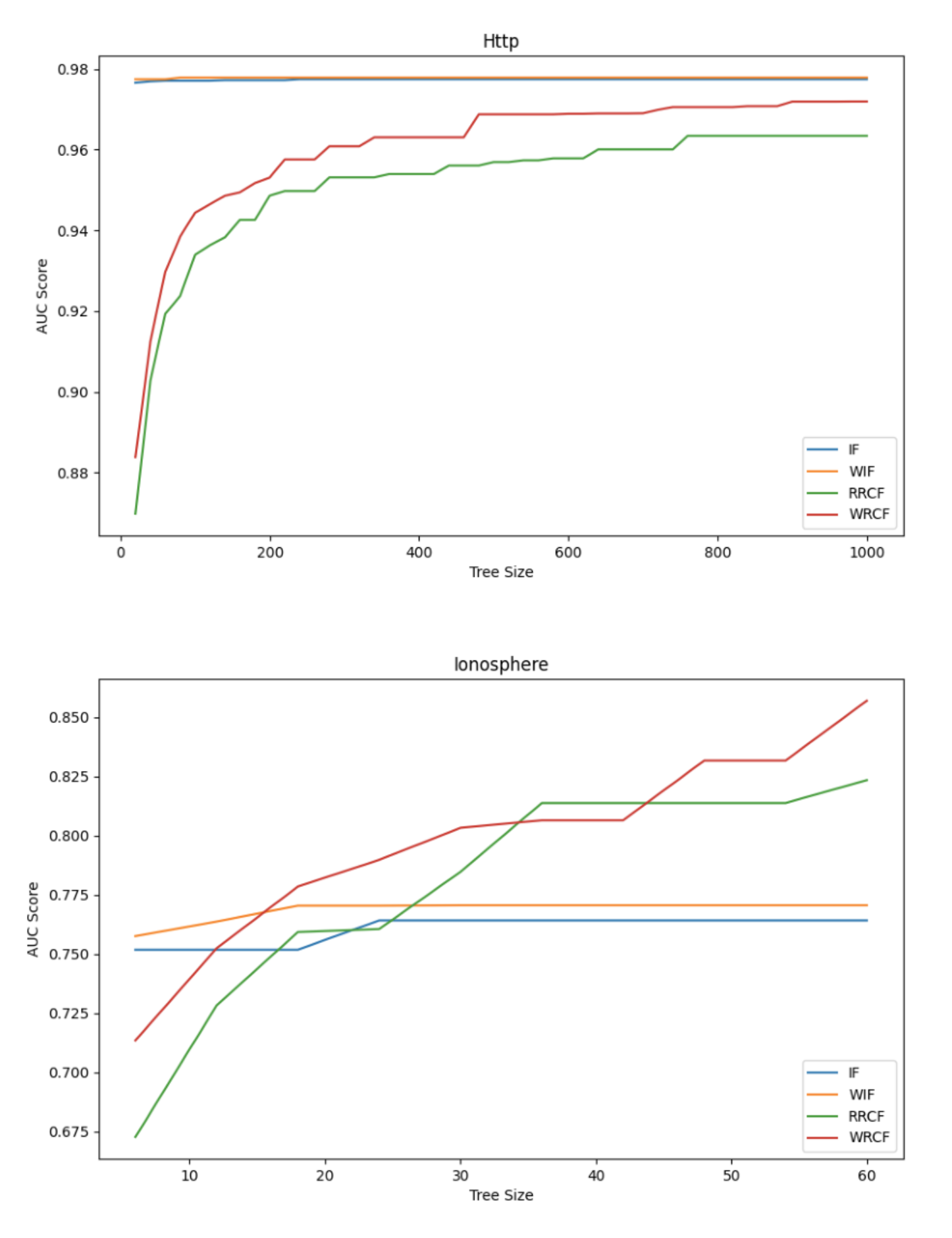}
    \includegraphics[scale=0.2]{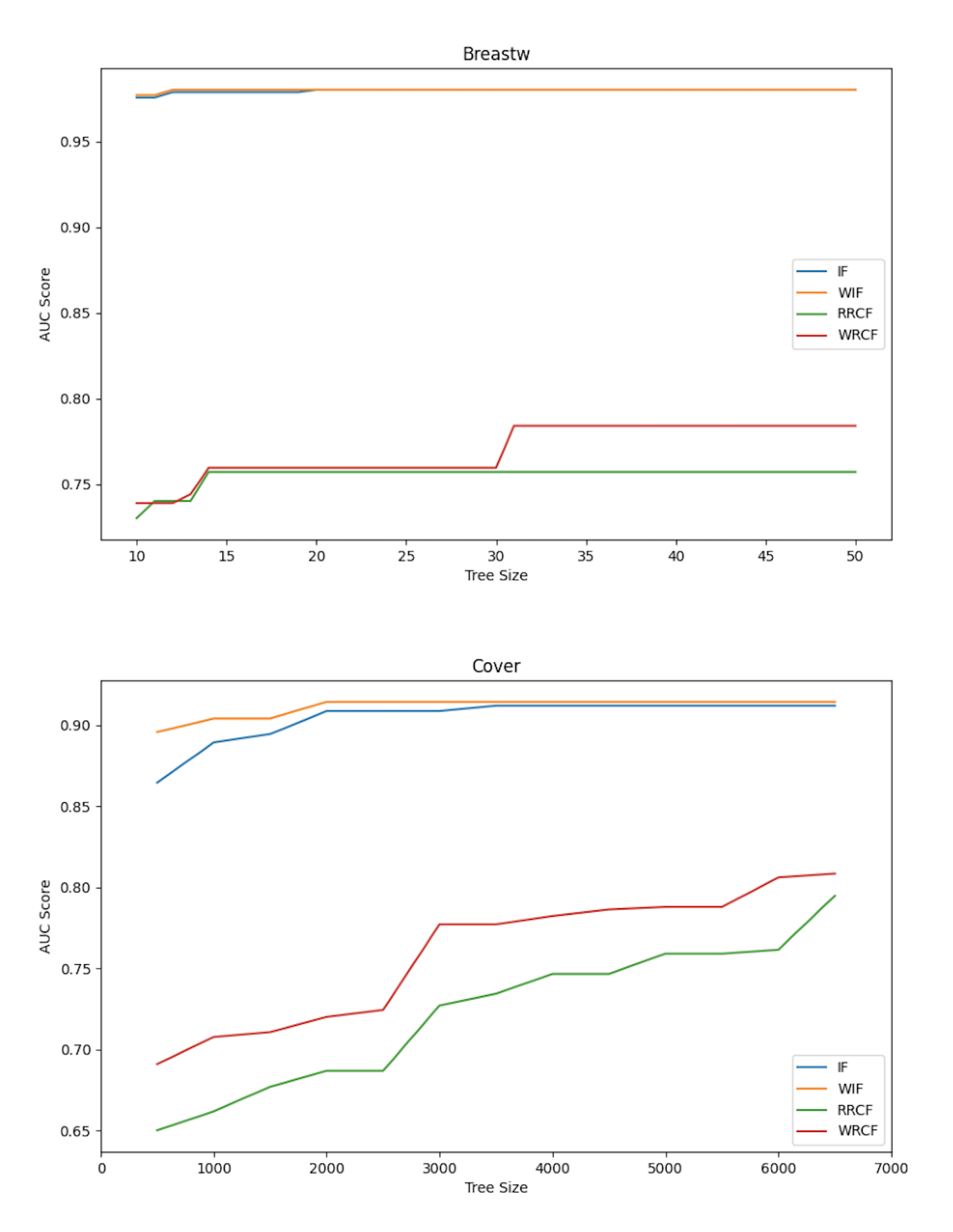}
    \caption{AUC scores versus tree size for Http and Inosphere (left column) and Breastw and Cover (right column)  data sets}
    \label{fig:Http and Inosphere}
\end{figure}

\begin{figure}[h] 
    \centering
    \includegraphics[scale=0.2]{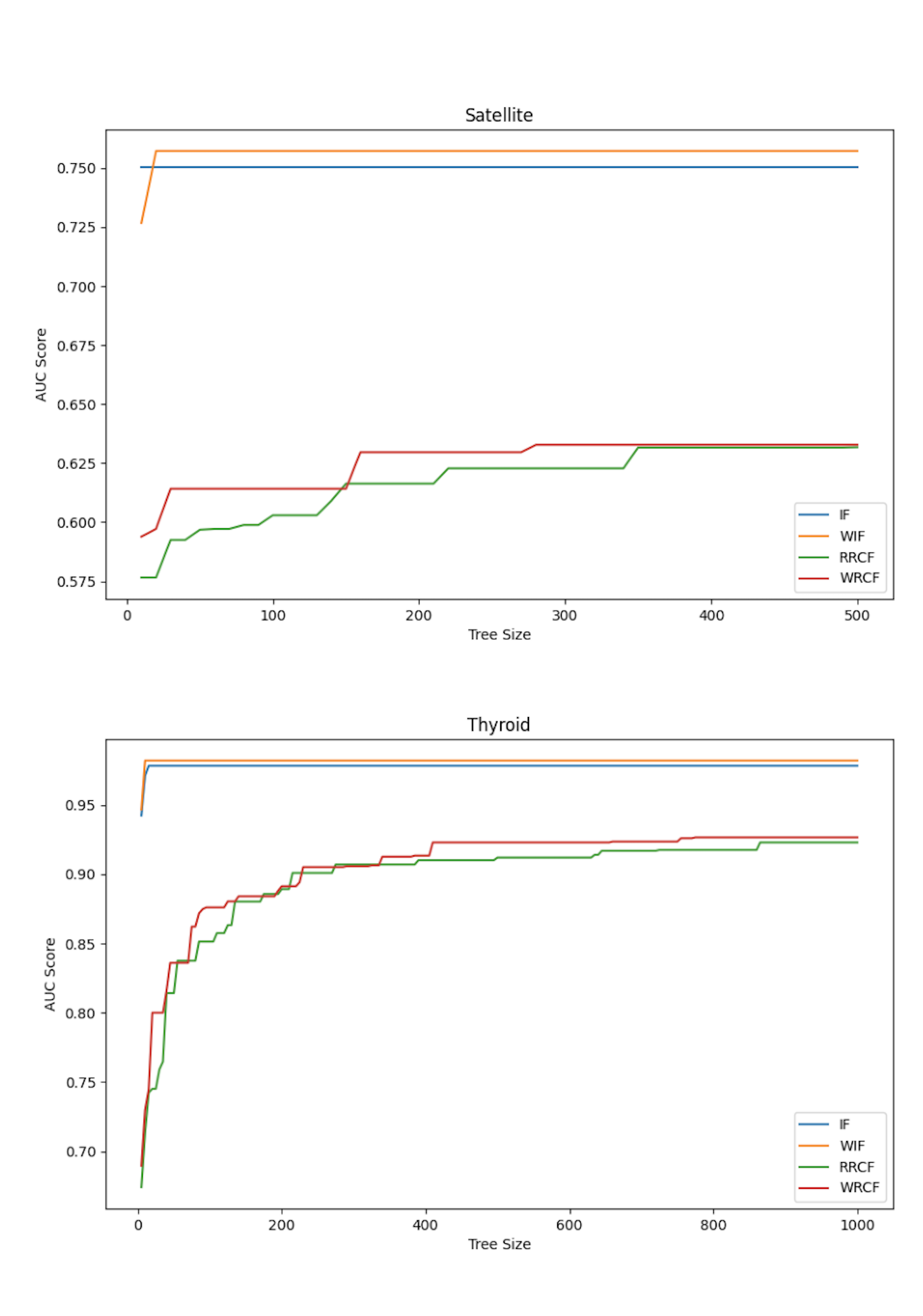}
        \includegraphics[scale=0.2]{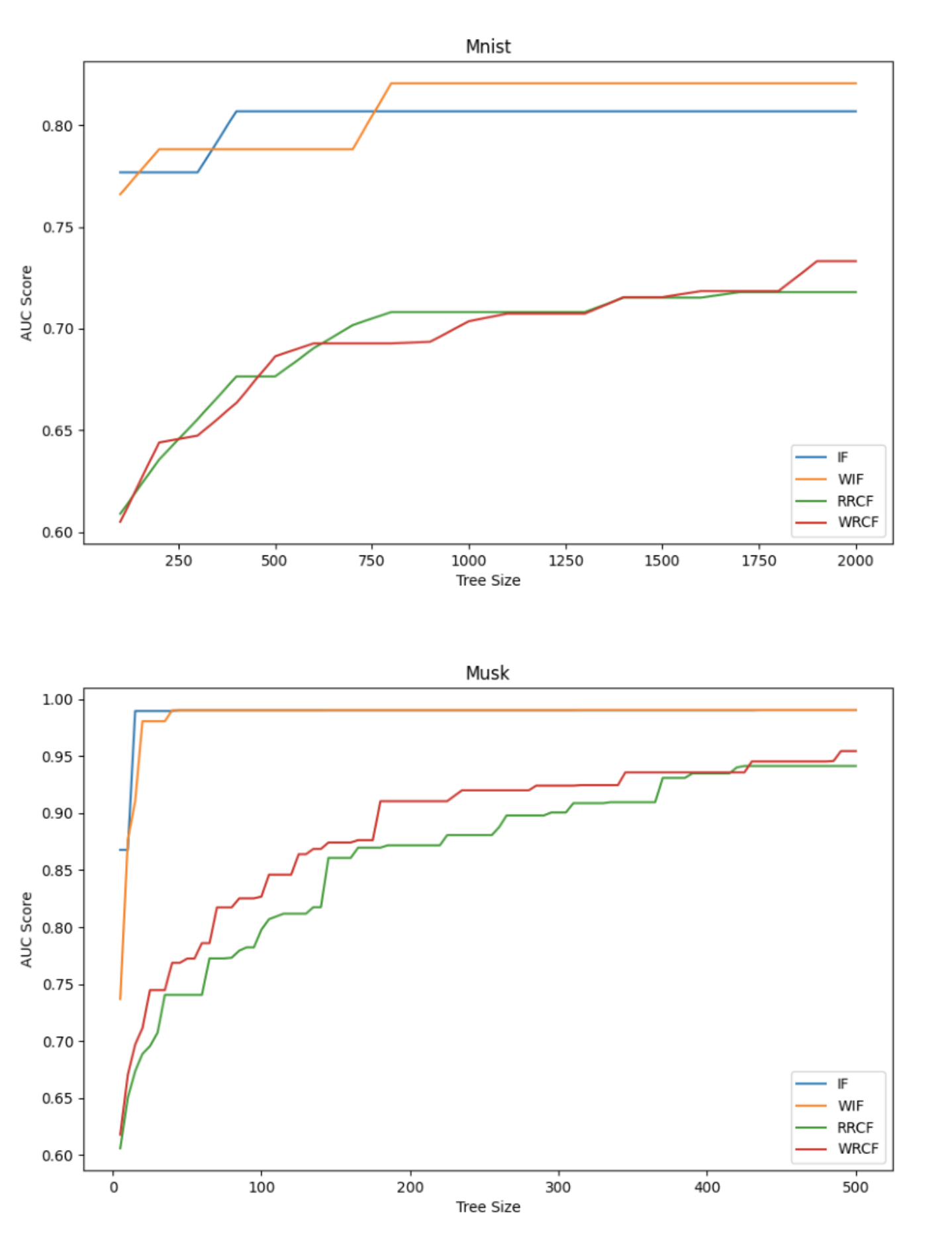}
    \caption{AUC scores versus tree size for Http and Inosphere (left column) and Breastw and Cover (right column)  data sets for Satellite and Thyroid (left column) and MNIST and Musk (right column) data sets}
    \label{fig:Satellite and Thyroid}
\end{figure}

Figures \ref{fig:Http and Inosphere} and \ref{fig:Satellite and Thyroid} show the results for each data set. The AUC scores are represented by the blue, orange, green and red colors for the IF, WIF, RRCF and WRCF, respectively. As the AUC score approaches 1, the algorithm demonstrates higher accuracy in binary classification. The results indicate that the WIF outperforms the IF, and the WRCF surpasses the RRCF across all evaluated data sets.

\begin{table}[h]
\centering
\begin{tabular}{|p{2cm}|p{1.3cm}|p{1.2cm}|p{5.6cm}|}
\hline
\textbf{Data set} & \textbf{Instances} & \textbf{Features} & \textbf{Description} \\
\hline
Http \cite{misc_kdd_cup_1999_data_130}  & 567,498 & 3 & The original data contains 41 attributes. However, it is reduced to 3 attributes, focusing on ‘http' service. \\
\hline
Ionosphere \cite{misc_ionosphere_52}  & 351 &33 & Binary classification of radar returns from the ionosphere.
 \\
\hline
Breastw \cite{misc_breast_cancer_wisconsin_(original)_15}  & 683& 9& Binary classification of breast cancer. \\
\hline
Cover \cite{misc_covertype_31}  & 286,048 &  10 & Classification of 7 forest cover types based on attributes such as elevation, aspect, and more. \\
\hline
Satellite \cite{misc_statlog_(landsat_satellite)_146} & 6,435
 & 36 & Multi-spectral values of pixels in 3x3 neighbourhoods in a satellite image, and the classification associated with the central pixel in each neighbourhood. \\
\hline
Thyroid \cite{misc_thyroid_disease_102} & 3,772& 6 & Classification data set for ANNs. \\
\hline
MNIST \cite{deng2012mnist} & 7,603 &100 & Handwritten digits data set transformed for outlier detection, treating digit-zero as inliers and digit-six as outliers. \\
\hline
Musk \cite{misc_musk_(version_2)_75} & 3,062 & 166& Binary classification of Musk data set. \\
\hline
\end{tabular}\caption{Benchmark data sets}
\label{table8}
\end{table}

%\begin{figure}[h]
%    \centering
%    \includegraphics[scale=0.45]{images/b2.png}
%    \caption{Breastw and Cover}
%    \label{fig:Breastw and Cover}
%\end{figure}

%\begin{figure}[h] 
%    \centering
%    \includegraphics[scale=0.45]{images/b4.png}
%    \caption{Mnist and Musk}
%    \label{fig:Mnist and Musk}
%\end{figure}

\FloatBarrier

\section{Conclusion}\label{Conclusion}
The IF and RCF  algorithms are efficient for anomaly detection. The RRCF algorithm, a variant of RCF, is efficient as it selects the dimension cuts based on the shape of data. In this paper, we showed that the IF and RRCF can be further improved using the density-aware partitioning when choosing the split values and proposed new algorithms, which we refer to as the weighted IF (WIF) and weighted RCF (WRCF) methods.
With the density-aware partitioning, the data structure is adaptively considered and faster convergence is obtained in determining the anomaly scores. To construct the WIF and WRCF, we proposed the concept of the density measure. We provided various mathematical properties of the density measure and density-aware partitioning. Numerical results provided in this paper verify that the proposed WIF and WRCF algorithms perform better than the IF and RRCF algorithms, respectively, especially when the data has non-uniformity in its structure.

\FloatBarrier

%\section*{Acknowledgment}
%This research was supported by National Research Foundation under the grant number 2021R1A2C3009648. JHJ was also supported partially by POSTECH Basic Science Research Institute under the grant number 2021R1A6A1A10042944. This work was also supported (in part) by the research grant from the NRF to the Center for the Gravitational-Wave Universe under the Grant Number 2021M3F7A1082053. 

\bibliography{sn-bibliography}
\end{document}